\documentclass[10pt,twocolumn,letterpaper]{article}
\usepackage{array}
\newcolumntype{L}{>{\centering\arraybackslash}m{3cm}}
\usepackage{cvpr}
\usepackage{times}
\usepackage{epsfig}
\usepackage{graphicx}
\usepackage{amsmath}
\usepackage{amssymb}

\DeclareMathOperator*{\argmax}{arg\,max}

 \makeatletter
 \DeclareRobustCommand\onedot{\futurelet\@let@token\@onedot}
 \def\@onedot{\ifx\@let@token.\else.\null\fi\xspace}
 \def\eg{e.g\onedot} 
 \def\ie{i.e\onedot}

 \makeatother

\DeclareRobustCommand{\secref}[1]{Section~\ref{#1}}

\newcommand{\mysubsubsection}[1]{\textbf{#1.}}
\newcommand{\mysubsection}[1]{\textbf{#1.}}

\usepackage[pagebackref=true,breaklinks=true,letterpaper=true,colorlinks,bookmarks=false]{hyperref}

\cvprfinalcopy 
\def\cvprPaperID{****} 

\ifcvprfinal\fi
\begin{document}

\title{Natural Language Object Retrieval
}

\author{Ronghang Hu$^1$ \quad Huazhe Xu$^2$ \quad Marcus Rohrbach$^{1,3}$ \quad Jiashi Feng$^4$ \quad Kate Saenko$^5$ \quad Trevor Darrell$^1$ \\
$^1$University of California, Berkeley \quad $^2$Tsinghua University \quad $^3$ICSI, Berkeley \\
$^4$National University of Singapore \quad $^5$University of Massachusetts, Lowell \\
{\tt\small \{ronghang, rohrbach, trevor\}@eecs.berkeley.edu, xhz12@mails.tsinghua.edu.cn} \\
{\tt\small elefjia@nus.edu.sg, saenko@cs.uml.edu }
}

\maketitle
\thispagestyle{empty}

\begin{abstract}
In this paper, we address the task of natural language object retrieval, to localize a target object within a given image based on a natural language query of the object. Natural language object retrieval differs from text-based image retrieval task as it involves spatial information about objects within the scene and global scene context. To address this issue, we propose a novel Spatial Context Recurrent ConvNet (SCRC) model as scoring function on candidate boxes for object retrieval, integrating spatial configurations and global scene-level contextual information into the network. Our model processes query text, local image descriptors, spatial configurations and global context features through a recurrent network, outputs the probability of the query text conditioned on each candidate box as a score for the box, and can transfer visual-linguistic knowledge from image captioning domain to our task. Experimental results demonstrate that our method effectively utilizes both local and global information, outperforming previous baseline methods significantly on different datasets and scenarios, and can exploit large scale vision and language datasets for knowledge transfer.
\end{abstract}

\section{Introduction}

\begin{figure}
\begin{center}
\includegraphics[width=0.90\linewidth]{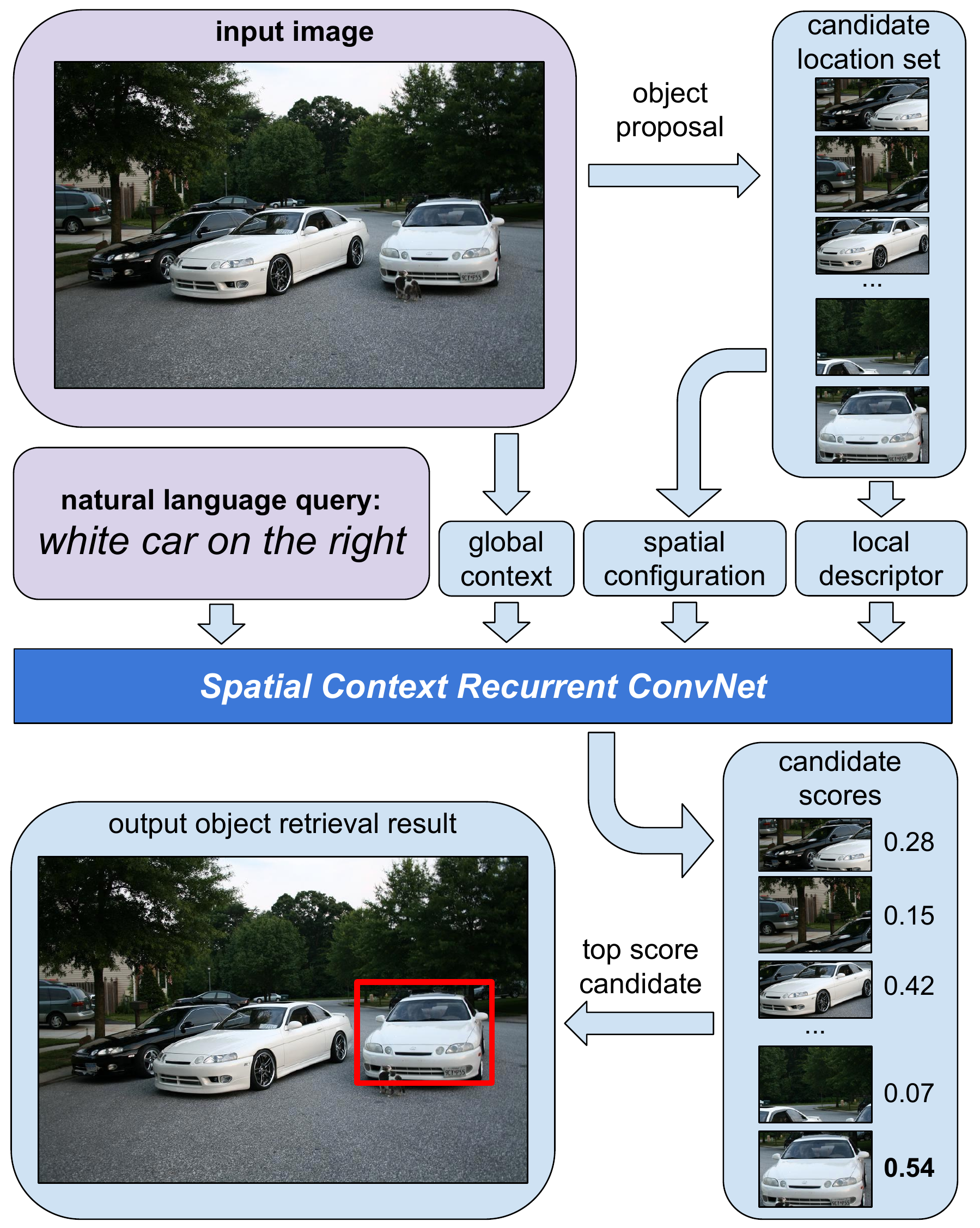}
\end{center}
\caption{Overview of our method. Given an input image, a text query and a set of candidate locations (\eg from object proposal methods), a recurrent neural network model is used to score candidate locations based on local descriptors, spatial configurations and global context. The highest scoring candidate is retrieved. }
\label{fig:teaser}
\end{figure}

Significant progress has been made in object detection in recent years; with the help of Convolutional Neural Networks (CNNs), it is possible to detect a predefined set of object categories with high accuracy \cite{girshick2014rich, girshick2015fast}, and the number of categories in object detection has grown over 10K to 100K with the help of domain adaptation \cite{hoffman2014lsda} and hashing \cite{dean2013fast}. However, in practical application scenarios, instead of using a predefined fixed set of object categories, one would often prefer to refer to an object with natural language rather than use a predefined category label. Such natural language query can include different types of phrases such as categories, attributes, spatial configurations and interactions with other objects, such as ``the young lady in a white dress sitting on the left'' or ``white car on the right'' in Figure \ref{fig:teaser}.

In this paper, we address the problem of natural language object retrieval: given an image and a natural language description of an object as query, we want to retrieve the object by localizing the object in the image. Natural language object retrieval can be seen as a generalization of generic object detection and has a wide range of applications, such as handling natural language commands in robotics where the user may ask to a robot to pick up ``the TV remote control on the shelf''.

We frame natural language object retrieval as a retrieval task on a set of candidate locations in a given image in this paper, as shown in Figure \ref{fig:teaser}, where candidate locations can come from object proposal methods such as EdgeBox \cite{zitnick2014edge}. We observe that simply applying text-based image retrieval systems on the image regions cropped from candidate locations for this task leads to inferior performance, as natural language object retrieval involves spatial configurations of objects and the global scene as context. For example, to decide how likely an object in a scene corresponds to ``the man in a blue jacket sitting on the right in front of the house'', one needs to look at both the object to determine whether it is ``the man'' (category), ``in blue jacket'' (attribute) and ``sitting'' (action), and its spatial configuration within the scene to determine whether it is ``on the right'', and the whole image as global contextual information to determine whether it is ``in front of the house''. Although both text-based image retrieval and natural language object retrieval involve jointly modeling images and text, they are different vision and language domains with domain shift from whole images to bounding boxes.

To address these issues, we propose the Spatial Context Recurrent ConvNet (SCRC) model to learn a scoring function 
that takes the text query, candidate regions, their spatial configurations and global context as input and outputs scores for candidate regions. Inspired by the Long-term Recurrent Convolutional Network (LRCN) \cite{donahue2015long}, an effective recurrent architecture for both image captioning and image retrieval, we use a two-layer LSTM network structure where the embedded text sequence and visual features serve as input to the first layer and the second layer, respectively. However, we note that it is possible to build our model on other recurrent network architectures such as \cite{mao2014explain, vinyals2015show}.

Compared with other types of visual-linguistic models such as bag-of-words \cite{ren2015image}, one of the advantages of using a recurrent neural network as scoring function is that the whole model can be easily learned end-to-end via simple back propagation, allowing visual feature extraction and text sequence embedding to be adapted to each other, and we show that it significantly outperforms a previous method using bag-of-words. Another advantage is that it is easy to utilize relatively large scale image-text datasets from other domains like image captioning (\eg MSCOCO \cite{lin2014microsoft}) to learn a vision-language model, by first pretraining the model on the image captioning task, and then adapting it to natural language object retrieval task through fine-tuning. One of the main challenges for natural language object retrieval is the lack of large scale datasets with annotated object bounding box and description pairs. To address this issue, we show that it allows us to transfer visual-linguistic knowledge learned from the former task to the latter one by first pretraining on the image caption domain and then adapting it to the natural language object retrieval domain. This pretraining and adaptation procedure improves the performance and avoids over-fitting, especially when the object retrieval training dataset is small.

\section{Related work}
\label{sec:related_work}

\begin{figure*}
\begin{center}
\includegraphics[width=0.90\linewidth]{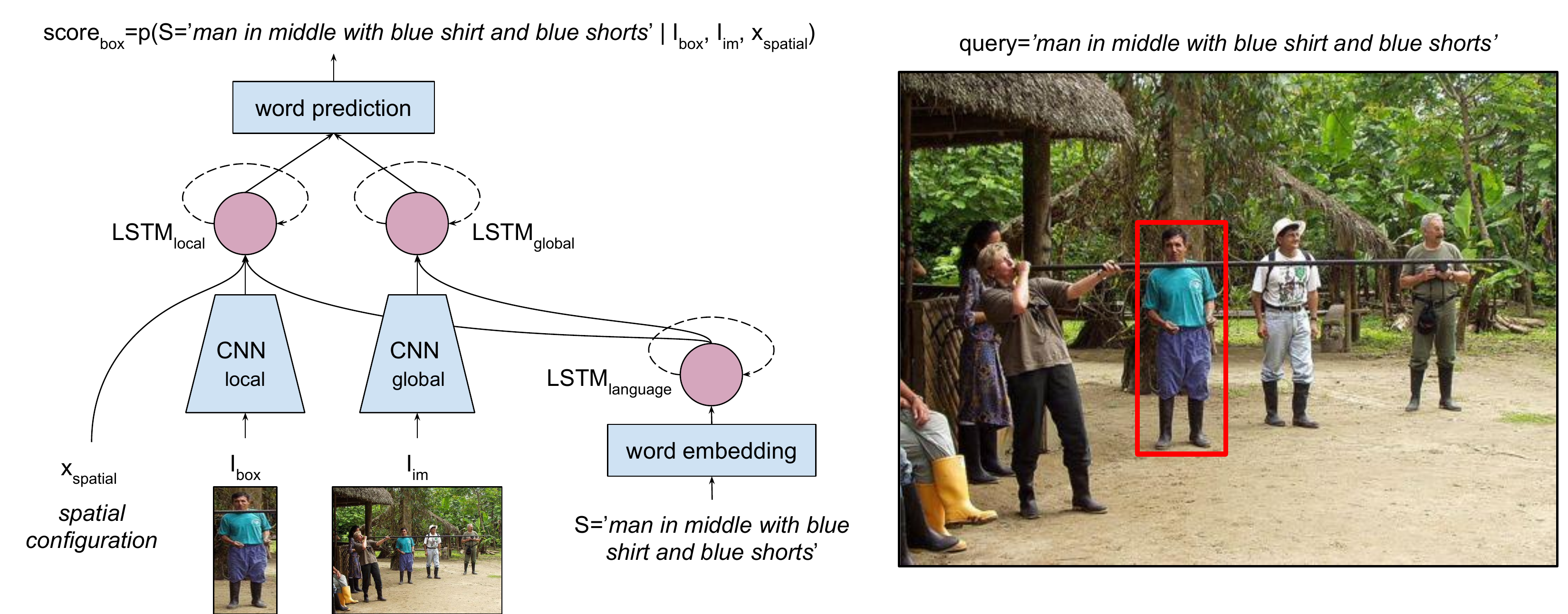}
\end{center}
\caption{Our Spatial Context Recurrent ConvNet (SCRC) for natural language object retrieval. The recurrent  network in our model contains three LSTM units. Two CNN's are used to extract local image descriptors and global scene-level contextual feature respectively. Parameters in word embedding, word prediction and three LSTM units are initialized by pretraining on image captioning dataset.}
\label{fig:method}
\end{figure*}

Natural language object retrieval, grounding, image captioning, and image retrieval   can be seen as different directions of the same super-task of jointly modeling a text sequence and image content, so it is natural to consider transferring knowledge learned from one task to another domain. In this work, we transfer knowledge from image captioning to natural language object retrieval by first pretraining our model on image captioning datasets to learn an initial parameter set for word embedding and word sequence prediction based on visual features. In the following we discuss these related areas.

\mysubsection{Natural language object retrieval}
Based on a bag of words sentence model and embeddings derived from ImageNET classifiers, \cite{guadarrama2014open} addresses a similar problem as ours and localizes an object within an image based on a text query. Given a set of candidate object regions, \cite{guadarrama2014open} generates text from those candidates represented as bag-of-words using category names predicted from a large scale pretrained classifier and compares the word bags to the query text. Other methods generate visual features from query text and match them to image regions, \eg through a text-based image search engine \cite{arandjelovic2012multiple} or learn a joint embedding of text phrases and visual features. Concurrent with our work, \cite{mao2015generation} also proposes a recurrent network model to localize objects from given descriptions.

\mysubsection{Grounding Objects from Image Descriptions}
Given an image and its description sentence, \cite{karpathy14nips} aligns sentence fragments to image regions by embedding the detection results from a pretrained object detector and the dependency tree from a parser with a ranking loss. \cite{karpathy15cvpr} builds on \cite{karpathy14nips} and replaces the dependency tree with a bidirectional RNN. Canonical Correlation Analysis (CCA) is used in \cite{plummer15iccv} to learn a joint embedding of image regions and text snippets to localize each object mentioned in the caption. \cite{kong2014you} uses a structure prediction model to align text to image and reasons about object co-reference in text for 3D scene parsing. Concurrent with this paper, \cite{rohrbach2015grounding} uses an attention model to ground referential phrases in image descriptions by attending to regions where the phrases can be best reconstructed.

\mysubsection{Image Captioning}
Image captioning methods take an input image and generate a text caption describing it. Recently, methods based on recurrent neural networks \cite{xu2015show, vinyals2015show, mao2014explain, donahue2015long} have shown to be effective on this task. LRCN \cite{donahue2015long} is one of these recent successful methods and involves a two-layer LSTM network with the embedded word sequence and image features as input at each time step. We use LRCN as our base network architecture in this work and incorporate spatial configurations and global context into the recurrent model for natural language object retrieval.

\mysubsection{Image Retrieval}
Text-based image retrieval systems select from a set of images an image that best matches the query text. In image retrieval, a ranking function is learned through a recurrent neural network \cite{mao2014explain, donahue2015long}, metric learning \cite{hoi2006learning}, correlation analysis \cite{klein2014fisher} and other methods \cite{frome2013devise, kiros2014unifying}. It was shown in \cite{donahue2015long} that a probabilistic image captioning model such as LRCN can also be used as an image retriever by using the probability of the query text sequence conditioned on the image $p( S_{query} | I )$ generated by image captioning model as a score for retrieval.

\section{Our model}
\label{sec:method}

In this section, we describe our Spatial Context Recurrent ConvNet (SCRC) model for natural language object retrieval and the training procedure in details. At test time, an image, a natural language object query and a set of candidate bounding boxes (\eg from object proposal methods such as EdgeBox \cite{zitnick2014edge}) are provided. The system needs to select from the candidate set a subset of bounding boxes that match the query text.

\subsection{Spatial Context Recurrent ConvNet}
\label{sec:the_network}

Inspired by the architecture of LRCN \cite{donahue2015long}, our Spatial Context Recurrent ConvNet (SCRC) model for natural language object retrieval consists of several components as illustrated in Figure \ref{fig:method}. The model has three Long Short-Term Memory (LSTM) \cite{hochreiter1997long} units denoted by $\text{LSTM}_{language}$, $\text{LSTM}_{local}$ and $\text{LSTM}_{global}$, a local and a global Convolutional Neural Network (CNN), a word embedding layer and a word prediction layer. At test time, given an image $I$, a query text sequence $S$ and a set of candidate bounding boxes $\{b_i\}$ in $I$, the network outputs a score $s_i$ for the $i$-th candidate box $b_i$ based on local image descriptors $x_{box}$ on $b_i$, spatial configuration $x_{spatial}$ of the box with respect to the scene, and global contextual feature $x_{context}$.

In this work, the local descriptor $x_{box}$ is extracted by $\text{CNN}_{local}$ from local region $I_{box}$ on $b_i$, and we use feature extracted by another network $\text{CNN}_{global}$ on the whole image $I_{im}$ as scene-level contextual feature $x_{context}$. The spatial configuration of $b_i$ is an 8-dimensional representation 
\begin{equation}
x_{spatial} = [ x_{\min}, y_{\min}, x_{\max}, y_{\max}, x_{\text{center}}, y_{\text{center}}, w_{box}, h_{box} ]
\label{eqn:spatial_config}
\end{equation} 
where $w_{box}$ and $h_{box}$ are the width and height of $b_i$. We normalize image height and width to be $2$ and place the origin at the image center, so that coordinates range from $-1$ to $1$.

The words $\{w_t\}$ in the query text sequence $S$ are represented as one-hot vectors and embedded through a linear word embedding matrix as $Ew_t$, and processed by $\text{LSTM}_{language}$ as the input time sequence. At each time step $t$, $\text{LSTM}_{local}$ takes in $[ h_{language}^{(t)}, x_{box}, x_{spatial} ]$ (concatenation of the three vectors, where $h_{language}^{(t)}$ is the hidden state from $\text{LSTM}_{language}$), and $\text{LSTM}_{global}$ takes in $[h_{language}^{(t)}, x_{context}]$. Finally, based on $h_{local}^{(t)}$ and $h_{global}^{(t)}$, a word prediction layer predicts the conditional probability distribution of the next word based on local image region $I_{box}$, whole image $I_{im}$, spatial configuration $x_{spatial}$ and all previous words it have seen so far, as
\begin{eqnarray}
& & p(w_{t+1} | w_t, \cdots, w_1, I_{box}, I_{im}, x_{spatial}) \nonumber \\
&=& \mathrm{Softmax} (W_{local} h_{local}^{(t)} + W_{global} h_{global}^{(t)} + r)
\label{eqn:next_word}
\end{eqnarray}
where $W_{local}$ and $W_{global}$ are weight matrices for word prediction and $r$ is a bias vector. $\mathrm{Softmax}(\cdot)$ is a softmax function over a vector to output a probability distribution.

We note that when setting $W_{local} = 0$ in Eqn. \ref{eqn:next_word}, our Spatial Context Recurrent ConvNet (SCRC) model is equivalent to the LRCN model \cite{donahue2015long} for image captioning and image retrieval by only modeling $p(S | I_{im} )$ to predict a text sequence $S$ based on the whole image $I_{im}$ while ignoring $I_{box}$ and $x_{spatial}$. This makes it possible to pretrain the model on the image captioning in Section \ref{sec:pretrain} to obtain a good parameter initialization for visual-linguistic modeling, and transfer knowledge from large image captioning datasets. 

We use VGG-16 net \cite{simonyan2014very} trained on ILSVRC-2012 dataset \cite{russakovsky2014imagenet} as the CNN architecture for $\text{CNN}_{local}$ and $\text{CNN}_{global}$ and extract 1000-dimensional \textit{fc8} outputs as $x_{box}$ and $x_{context}$, and use the same LSTM implementation as in \cite{donahue2015long}, where the gates are computed as
\begin{eqnarray}
i_t &=& \sigma(W_{xi} x_t + W_{hi} h_{t-1} + b_i) \label{eqn:lstm_1} \\
f_t &=& \sigma(W_{xf} x_t + W_{hf} h_{t-1} + b_f) \\
o_t &=& \sigma(W_{xo} x_t + W_{ho} h_{t-1} + b_o) \\
g_t &=& \tanh(W_{xg} x_t + W_{hg} h_{t-1} + b_g)  \label{eqn:lstm_4}
\end{eqnarray}
All the three LSTM units have 1000-dimensional state $h_t$.

At test time, given an input image $I$, a query text $S$ and a set of candidate bounding boxes $\{b_i\}$, the query text $S$ is scored on $i$-th candidate box using the likelihood of $S$ conditioned on the local image region, the whole image and the spatial configuration of the box, computed as
\begin{eqnarray}
s &=& p(S | I_{box}, I_{im}, x_{spatial}) \nonumber \\
&=& \prod_{w_t \in S} p(w_t | w_{t-1}, \cdots, w_1, I_{box}, I_{im}, x_{spatial}) \label{eqn:joint_prob}
\end{eqnarray}
and the highest scoring candidate boxes are retrieved.

\subsection{Knowledge transfer from image captioning}
\label{sec:pretrain}

To exploit paired image-text data in image captioning datasets, and to learn a good initialization of parameters in word embedding, word prediction and three LSTM units, we first pretrain our model on an image captioning dataset, by restricting $W_{local} = 0$ in Eqn. \ref{eqn:next_word}, which is equivalent to training a LRCN model \cite{donahue2015long}. We follow the procedure in \cite{donahue2015long} for pretraining on image captioning. During pretraining, the probability of ground truth image caption $p(S_{gt} | I_{im})$ is maximized over the training image-sentence pairs, and the whole network is optimized with standard Stochastic Gradient Descent (SGD). We refer to \cite{donahue2015long} for the training details on image captioning.

Since we restrict $W_{local} = 0$ in Eqn. \ref{eqn:next_word} during pretraining, the parameters in $\text{LSTM}_{local}$ are not learned. To obtain a good initialization of this unit, we copy those weights in Eqn. \ref{eqn:lstm_1} -- \ref{eqn:lstm_4} from $\text{LSTM}_{global}$ to $\text{LSTM}_{local}$. The weights over the extra 8 dimensions of $x_{spatial}$ are initialized with zero. We also copy $W_{global}$ to $W_{local}$ to initialize word prediction weights.

After pretraining on the image captioning task, the parameters in our model already encode useful knowledge of word embedding and decoding and sequence prediction based on image features. The knowledge is transferred to the natural language object retrieval task in Section \ref{sec:adaptation}.

\subsection{Training for object retrieval}
\label{sec:adaptation}

After pretraining, we adapt the SCRC model to natural language object retrieval. In this paper, we assume that the training dataset consists of $N$ images, with each image containing $M_i$ ($i = 1, \cdots, N$) annotated objects, and each object annotated by a bounding box and $K_{i,j}$ ($i = 1, \cdots, N$, $j = 1, \cdots, M_i$) text descriptions (an object can be described more than once with different descriptions). At training time, each instance is an image-bounding box-description tuple $(I_{i}, b_{i,j}, S_{i, j, k})$, where $I_i$ is the whole image, $b_{i,j}=[x_{\min}, y_{\min}, x_{\max}, y_{\max}]$ is the bounding box of the $j$-th object and $S_{i, j, k}$ is a description text in natural language such as ``the black and white cat''. 

Our model for natural language object retrieval can be trained via maximizing the probability of the object description text in ground truth annotations conditioned on the local image region $I_{box}$ and the whole image $I_{im}$ as context, which is analogous to training a generic object detection system. Many state-of-the-art generic object detectors \cite{girshick2014rich, girshick2015fast} are built by turning object detection into a classification problem on candidate bounding boxes produced either from a sliding window or an object proposal mechanism, and a classifier is trained by maximizing the probability of ground truth object category label. In natural language object retrieval, the description text of an object can be seen as a generalized ``label'' of the object, and maximizing its conditional probability is similar to training a ``generalized classifier'' whose output is a sequence of word labels rather than a single category label.

Given a natural language object retrieval dataset, we construct all tuples $(I_{i}, b_{i,j}, S_{i, j, k})$ from the ground truth annotations as training instances (multiple tuples are constructed if there are multiple descriptions for the same object). For each annotated object in the training set, an image patch $I_{box}$ is cropped from the whole image $I_{im}$ based on bounding box of that object region, with its spatial configuration $x_{spatial}$ constructed through Eqn. \ref{eqn:spatial_config}. We define the loss function during training as
\begin{equation}
L = - \sum_{i=1}^N \sum_{j=1}^{M_i} \sum_{k=1}^{K_{i,j}} \log(p(S_{i,j,k} | {I_{box}}_{i,j}, {I_{im}}_{i}, {x_{spatial}}_{i,j}))
\end{equation}
where $N$ is the number of images, $M_i$ is the number of annotated objects in $i$-th image, $K_{i,j}$ is the number of natural language descriptions associated with the $j$-th object in that image, and $p(S_{i,j,k} | {I_{box}}_{i,j}, {I_{im}}_{i}, {x_{spatial}}_{i,j})$ is computed by Eqn. \ref{eqn:joint_prob}.

During training, the model parameters are initialized from the pretrained network in Section \ref{sec:pretrain}, and fine-tuned using SGD with a smaller learning rate, allowing the network to adapt to natural language object retrieval domain. The whole network is trained end-to-end via back propagation. Our model is implemented using Caffe \cite{jia2014caffe} and our code and data are available at \url{http://ronghanghu.com/text_obj_retrieval}.

\section{Experiments}
\label{sec:exp}

We evaluate our method on different datasets from small scale to relatively large scale. Since  \cite{guadarrama2014open} solves a similar problem to our paper, we adopt it as our baseline. In \cite{guadarrama2014open}, a large scale fine-grained classifier of 7K object classes is trained on ImageNET \cite{deng2009imagenet}. Each box in the candidate set is classified into one of the 7K classes, and a bag of words is extracted from the predicted object class based on its ImageNET \cite{deng2009imagenet} synset containing category name and synonyms. Then, the word bag is projected to a vector space, and matched to the projected query text using cosine distance to obtain a score. The sentence projection (embedding) in \cite{guadarrama2014open} is predefined and the only training involved in training the 7K object classifier. Note that \cite{guadarrama2014open} also proposes an instance match model that relies on online APIs at test time. As in this work we assume a self-contained system without resorting to other APIs, we only use the category model (CAFFE-7K) in \cite{guadarrama2014open} as our baseline.

As our recurrent architecture is inspired by LRCN \cite{donahue2015long}, which is shown to be effective for both image captioning and image retrieval, we also compare our model to LRCN. We use the LRCN model trained on MSCOCO \cite{lin2014microsoft} for image captioning task as an object retriever by evaluating it on candidate bounding boxes. Given an image $I$ with a set of candidate boxes and a query text $S_{query}$, we compute $p(S_{query} | I_{box})$, the probability of the query text $S_{query}$ conditioned on the local image region $I_{box}$ outputted by LRCN as a score for each box in the candidate set, and retrieve highest scoring candidates.

\subsection{Object retrieval evaluation on ReferIt dataset}
\label{sec:exp_referit}

The ReferIt dataset \cite{kazemzadeh2014referitgame} is the biggest publicly available dataset containing image regions with descriptions at the time of writing. It contains 20,000 images from IAPR TC-12 dataset \cite{grubinger2006iapr}, together with segmented image regions from SAIAPR-12 dataset \cite{escalante2010segmented}, and 120K annotated descriptions for the image regions collected in a two-player game that aims to make the image region identifiable from the annotation. The ReferIt dataset also contains some ambiguous (\eg ``anywhere'') and mistakenly annotated examples where the annotation does not correspond to any object. To evaluate on this dataset, we split the 20,000 images (together with their annotations) into 10,000 for training and validation and 10,000 for test, and construct image-bounding box-description tuples on all annotated image regions as training instances. There are 59,976 \textit{(image, bounding box, description)} tuples in the trainval set and 60,105 in the test set. In our experiments on this dataset, we only use the bounding boxes of annotated regions during training and evaluation. The bounding boxes are obtained from the segmentation regions in SAIAPR-12 dataset corresponding to the clicks by annotators. Note that although \cite{kazemzadeh2014referitgame} introduces the ReferIt dataset, it does not propose a baseline method for object retrieval based on text query.

As described in Section \ref{sec:method}, we first pretrain a SCRC model on MSCOCO dataset \cite{lin2014microsoft} for image captioning. The training details such as hyper-parameters of SGD follow \cite{donahue2015long}. After pretraining, we copy the weights in LSTM and the word prediction layer to the local part of the network as mentioned in Section \ref{sec:pretrain}. Then the pretrained SCRC model is adapted to the natural language object retrieval task following the procedure in Section \ref{sec:adaptation}. The model is fine-tuned on image-bounding box-description tuples in ReferIt trainval set with back propagation.

\mysubsubsection{Ablations}
To test the effect of incorporating spatial configurations $x_{spatial}$ and scene-level contextual feature $x_{context}$, we evaluate different setups during fine-tuning on ReferIt. By setting $x_{spatial}$ and $W_{global}$ to $0$ during fine-tuning and testing, the model can only learn to score a box based on local image descriptors $x_{box}$ from candidate boxes, denoted by \textbf{SCRC (w/o context, spatial)}. Similarly, by setting $W_{global}$ to $0$, the model can learn a scoring function on $x_{box}$ and $x_{spatial}$ but cannot utilize scene-level context, denoted by \textbf{SCRC (w/o context)}.

As a comparison, we directly trained a SCRC model on ReferIt without first pretraining on MSCOCO, and set $x_{spatial}$ and $W_{global}$ to $0$ during training and testing, denoted by \textbf{SCRC (w/o context, spatial, transfer)}. The CNN parameters in the model are initialized from VGG-16 net \cite{simonyan2014very} and other parameters are randomly initialized. In all the training above, the whole SCRC model is trained end-to-end with SGD, allowing visual feature extraction and textual sequence prediction to be optimized jointly.

At test time, all the 4 SCRC models mentioned above, the bag-of-words model (CAFFE-7K) in \cite{guadarrama2014open} and LRCN \cite{donahue2015long} as an object retriever on candidate boxes are compared on the ReferIt test set. The LRCN model is trained on MSCOCO dataset for image captioning as described in \cite{donahue2015long} to learn a probabilistic generative model $p(S|I)$, and we use it to score a candidate region $I_{box}$ based on a text query $S_{query}$ by computing the probability of the text conditioned on the local region, \ie $p(S_{query}|I_{box})$ as a baseline.

We evaluate with two testing scenarios: In the first scenario similar to the experiment in \cite{guadarrama2014open}, given an image and a text query, the model is asked to retrieve the corresponding image region from all annotated regions in that image. In the second scenario, which is a harder task but closer to real applications, given a text query the model retrieves an image region from a set of candidate bounding boxes produced by object proposal methods. A retrieved region is considered as correct if it overlaps with ground truth bounding box by at least 50\% IoU. In this experiment, we use top 100 proposals from EdgeBox \cite{zitnick2014edge} as our candidate box set.

\begin{table}[t]
\begin{center}
\small
\begin{tabular}{|l|c|c|}
\hline
\textbf{Method} & P@1-NR & P@1 \\
\hline
CAFFE-7K \cite{guadarrama2014open}          &  32.53\% &  27.73\%           \\
LRCN \cite{donahue2015long}                 &     -    &  38.38\%           \\
SCRC (w/o context, spatial, transfer)       &     -    &  61.03\%           \\
SCRC (w/o context, spatial)                 &     -    &  64.09\%           \\
SCRC (w/o context)                          &     -    &  70.15\%           \\
SCRC                                        &     -    &  \textbf{72.74\%}  \\
\hline
\end{tabular}
\end{center}
\caption{Top-1 precision of our method compared with baselines on annotated bounding boxes in ReferIt dataset. See \secref{sec:referit:results} for details.}
\label{tab:results_referit_gt}
\end{table}

\mysubsubsection{Results}
\label{sec:referit:results}
Table \ref{tab:results_referit_gt} shows the top-1 precision (the percentage of the highest-scoring region being correct) in the first scenario where the candidate set is all annotated boxes in the image. Note that CAFFE-7K cannot return informative results when none of the words in query are in its category names (leading to an empty bag and same score for all regions), whereas our SCRC model can always return deterministic result since it can represent unknown words with ``\textless unk\textgreater''. Similar to \cite{guadarrama2014open}, we evaluate with ``P@1-NR'' corresponding to non-random top-1 precision computed on the those informative results and ``P@1'' corresponding to top-1 precision on all cases including non-informative results, where random guess is used.  Results show that our full SCRC model achieves the highest top-1 precision. In Table \ref{tab:results_referit_gt}, it can be seen that pretraining on image captioning, adding spatial configuration, and adding scene-level context all improve the performance, with adding spatial configuration $x_{spatial}$ leading to the most significant performance boost. This is no surprise, as spatial configuration not only benefits in cases where spatial relationship is directly involved in the query (\eg ``the man on the left''), but also enables the network to learn a prior distribution of object locations (\eg ``sky'' is usually at the top of the scene while ``ground'' is usually at the bottom). 

\begin{figure*}
\begin{center}
\begin{tabular}{c@{}c@{}c}

\small{query='\textit{man squatting}'} & \small{query='\textit{standing guy}'} & \small{query='\textit{bike wheels}'} \\
\includegraphics[trim = 40mm 25mm 40mm 20mm , clip=true,width=0.3\textwidth,height=\textheight,keepaspectratio]{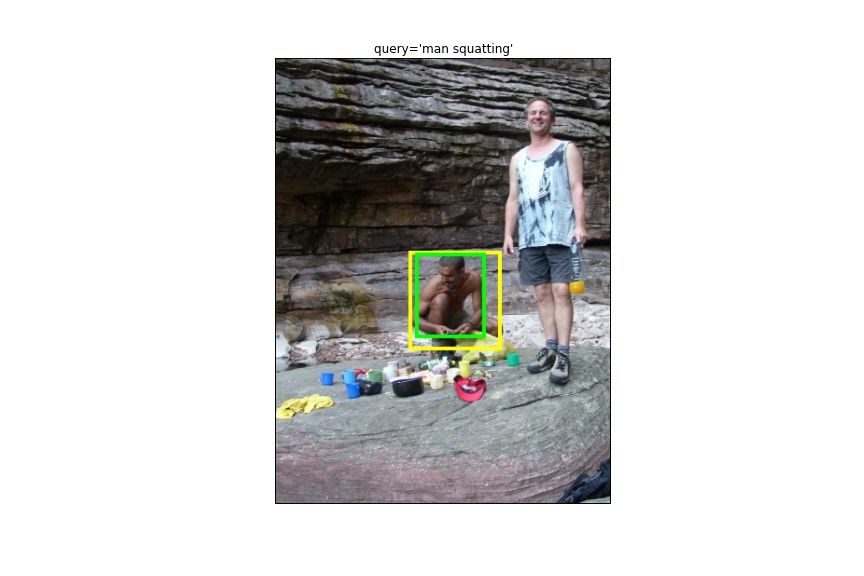} & 
\includegraphics[trim = 40mm 25mm 40mm 20mm , clip=true,width=0.3\textwidth,height=\textheight,keepaspectratio]{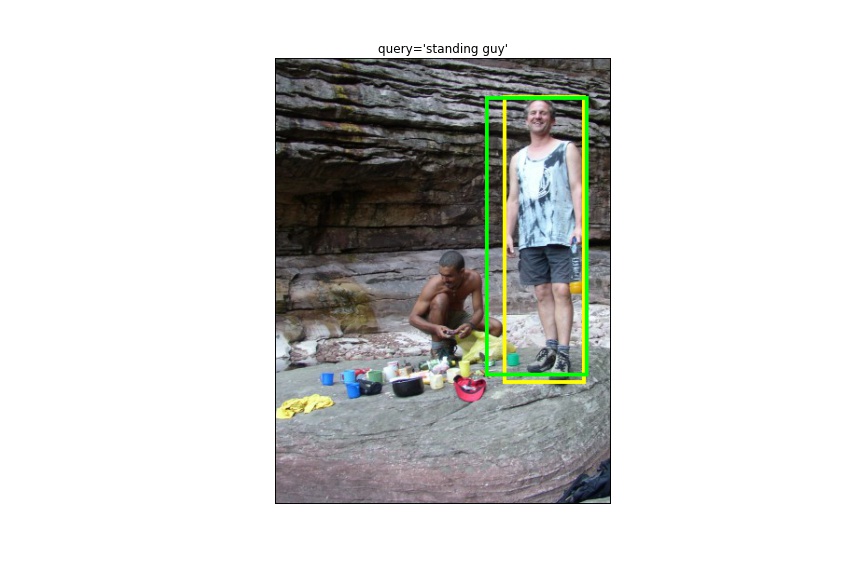} &
\includegraphics[trim = 40mm 25mm 40mm 20mm , clip=true,width=0.3\textwidth,height=\textheight,keepaspectratio]{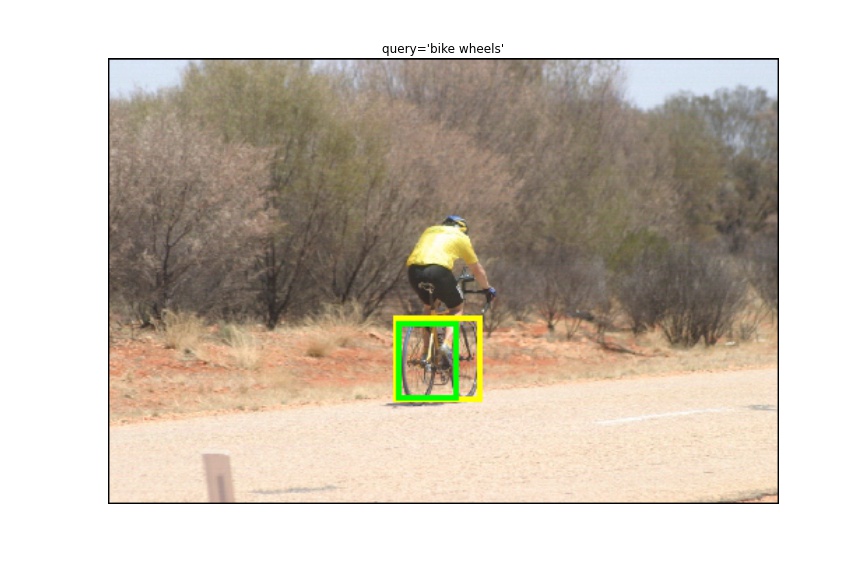} \\

\small{query='\textit{white hat}'} & \small{query='\textit{Window with closed curtains}'} & \small{query='\textit{right lake}'} \\
\includegraphics[trim = 40mm 25mm 40mm 20mm , clip=true,width=0.3\textwidth,height=\textheight,keepaspectratio]{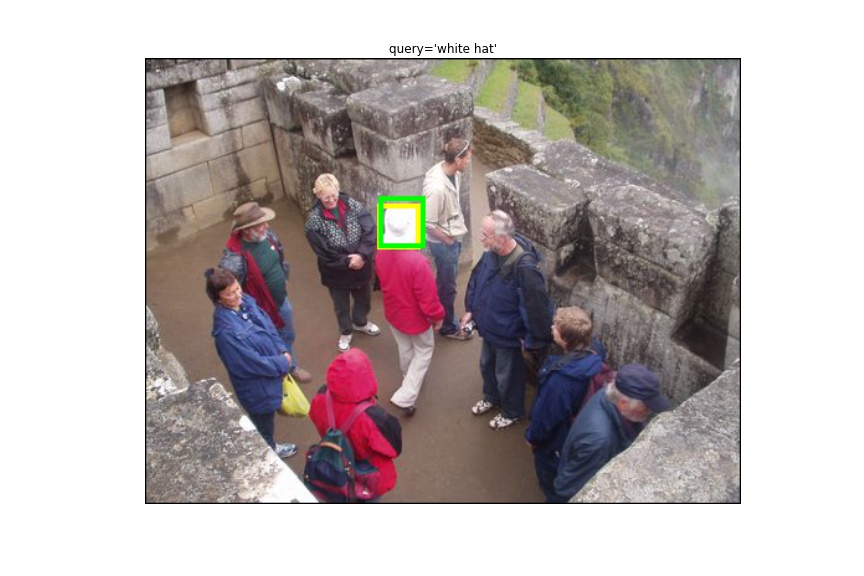} & 
\includegraphics[trim = 40mm 25mm 40mm 20mm , clip=true,width=0.3\textwidth,height=\textheight,keepaspectratio]{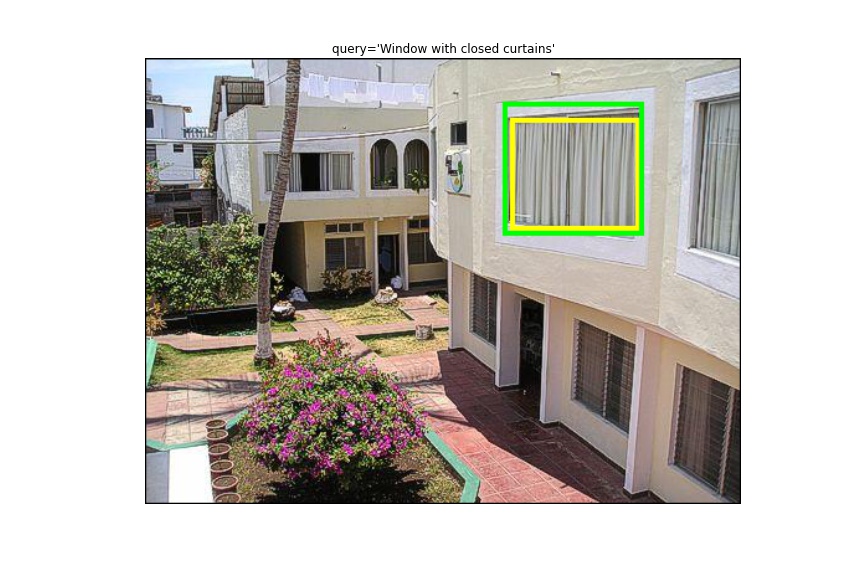} &
\includegraphics[trim = 40mm 25mm 40mm 20mm , clip=true,width=0.3\textwidth,height=\textheight,keepaspectratio]{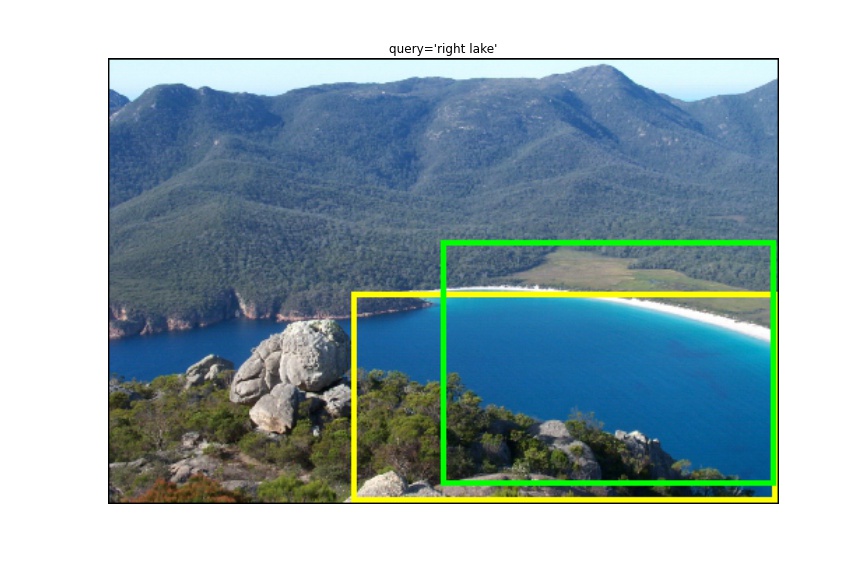} \\

\small{query='\textit{bird on the left}'} & \small{query='\textit{leaves of left tree}'} & \small{query='\textit{pillar building in the middle}'} \\
\includegraphics[trim = 40mm 25mm 40mm 20mm , clip=true,width=0.3\textwidth,height=\textheight,keepaspectratio]{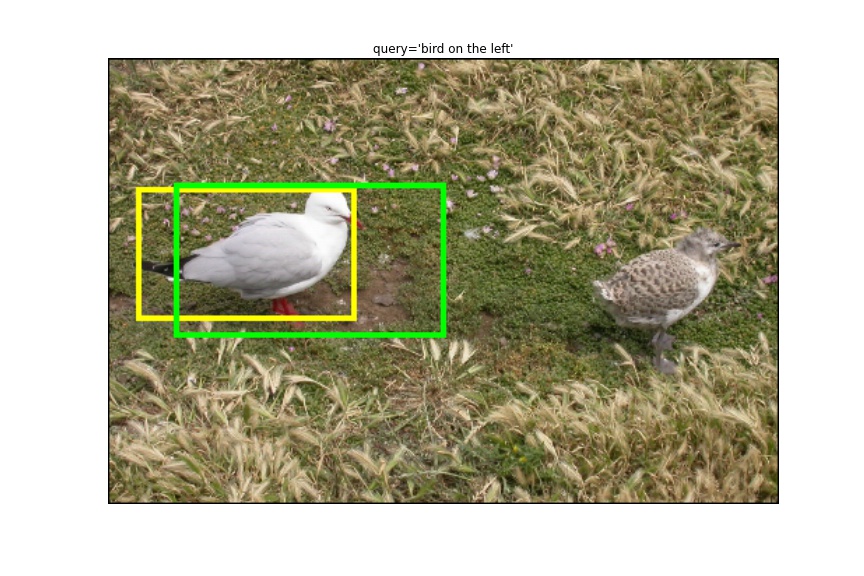} & 
\includegraphics[trim = 40mm 25mm 40mm 20mm , clip=true,width=0.3\textwidth,height=\textheight,keepaspectratio]{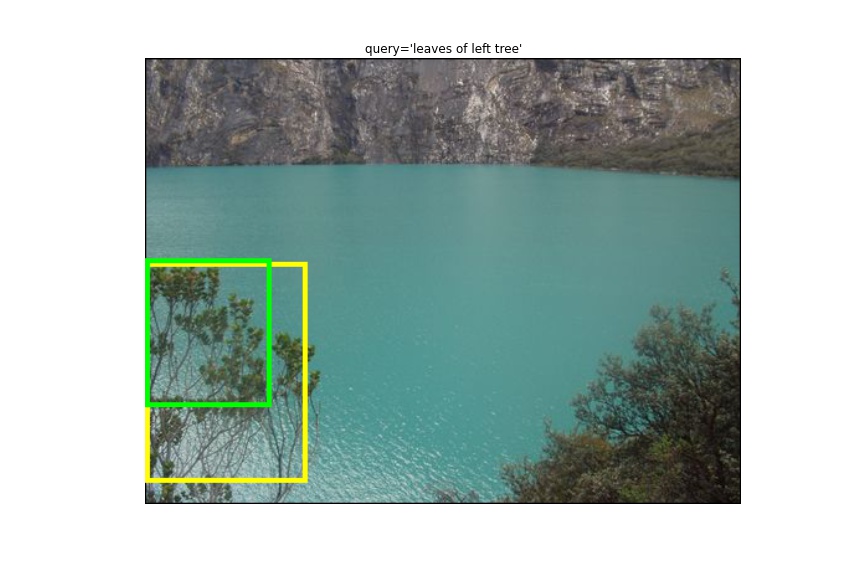} &
\includegraphics[trim = 40mm 25mm 40mm 20mm , clip=true,width=0.3\textwidth,height=\textheight,keepaspectratio]{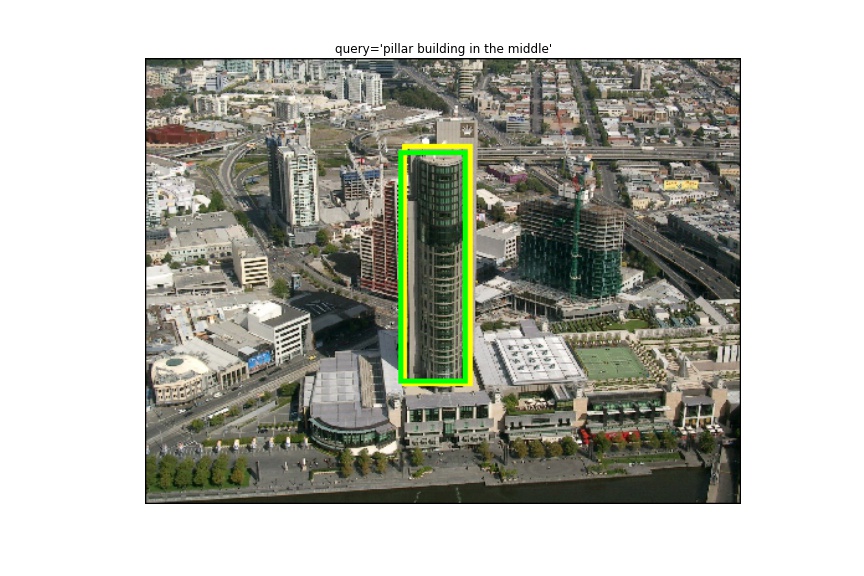}
\end{tabular}
\end{center}
\caption{Correctly localized examples ($\mathrm{IoU} \geq 0.5$) on ReferIt with EdgeBox. Ground truth in yellow and correctly retrieved box in green.}
\label{fig:referit_success}

\begin{center}
\begin{tabular}{c@{}c@{}c}

\small{query='\textit{man on right blue gloves}'} & \small{query='\textit{the piece with no shadows}'} & \small{query='\textit{face}'} \\
\includegraphics[trim = 40mm 25mm 40mm 20mm , clip=true,width=0.3\textwidth,height=\textheight,keepaspectratio]{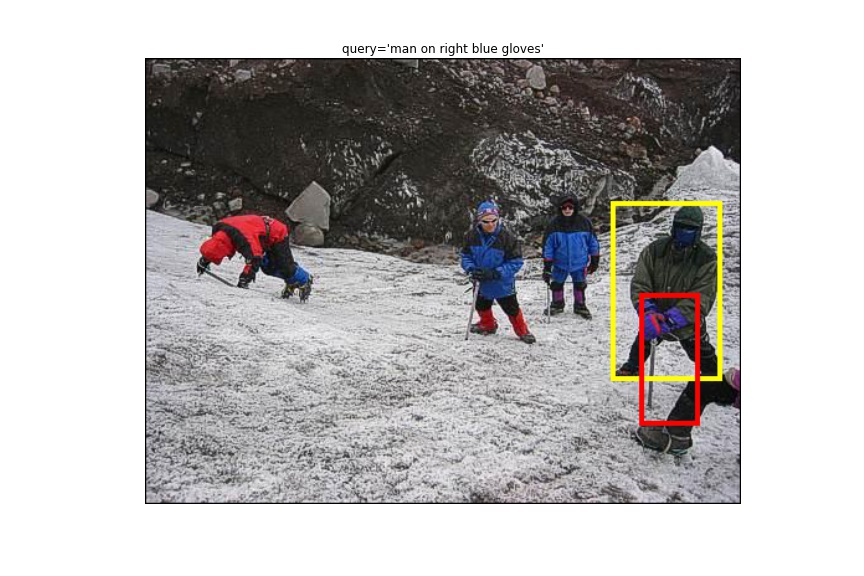} & 
\includegraphics[trim = 40mm 25mm 40mm 20mm , clip=true,width=0.3\textwidth,height=\textheight,keepaspectratio]{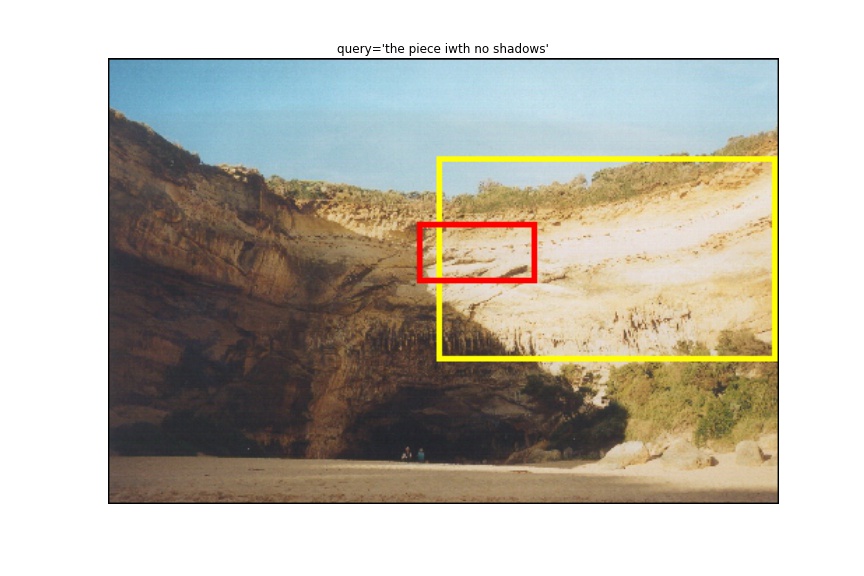} & 
\includegraphics[trim = 40mm 25mm 40mm 20mm , clip=true,width=0.3\textwidth,height=\textheight,keepaspectratio]{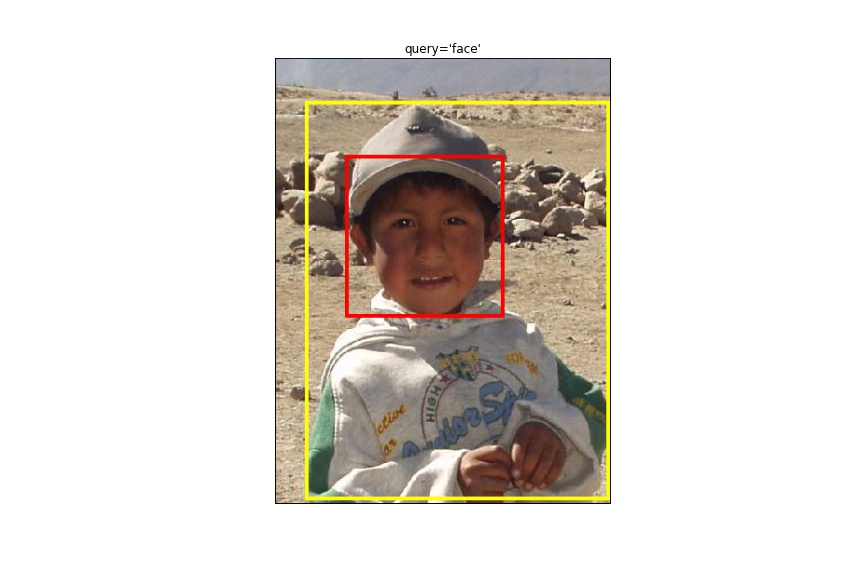} \\

\small{query='\textit{rock}'} & \small{query='\textit{water on the right only}'} & \small{query='\textit{dirt patch next to car (right side)}'} \\
\includegraphics[trim = 40mm 25mm 40mm 20mm , clip=true,width=0.3\textwidth,height=\textheight,keepaspectratio]{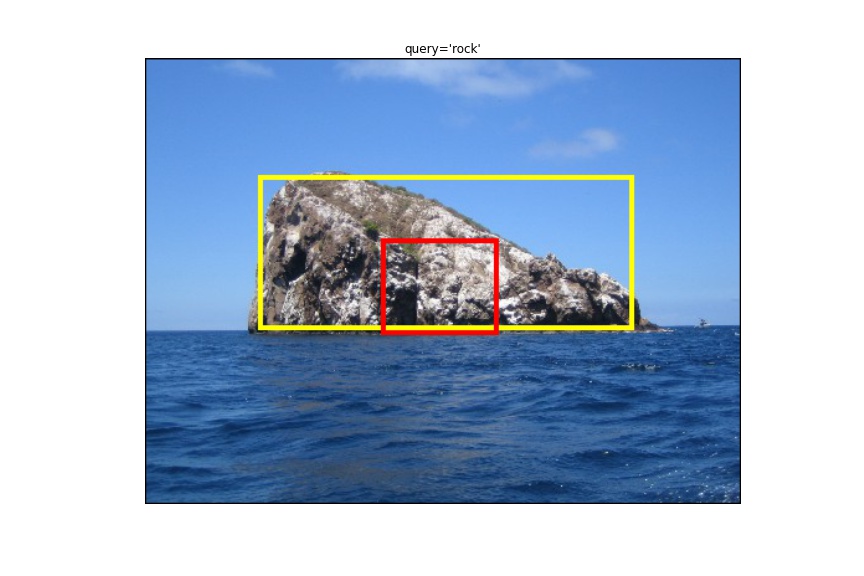} & 
\includegraphics[trim = 40mm 25mm 40mm 20mm , clip=true,width=0.3\textwidth,height=\textheight,keepaspectratio]{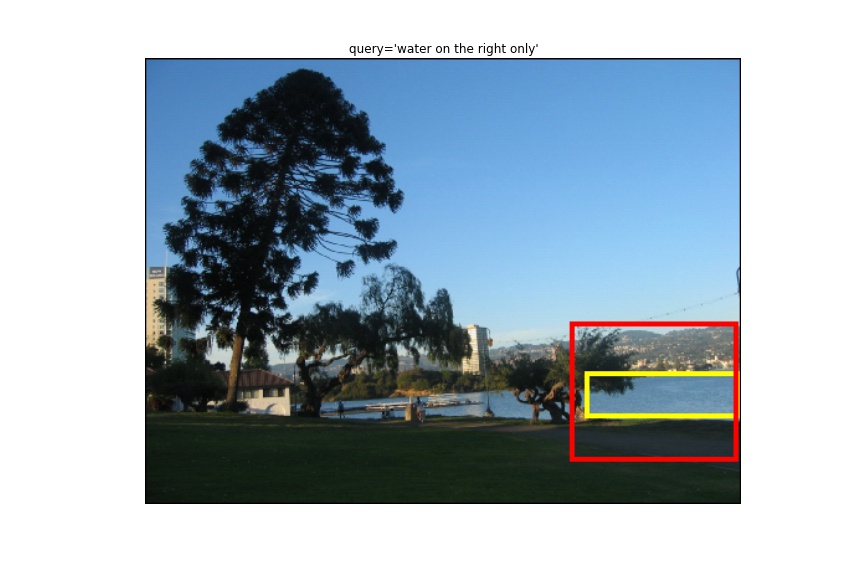} & 
\includegraphics[trim = 40mm 25mm 40mm 20mm , clip=true,width=0.3\textwidth,height=\textheight,keepaspectratio]{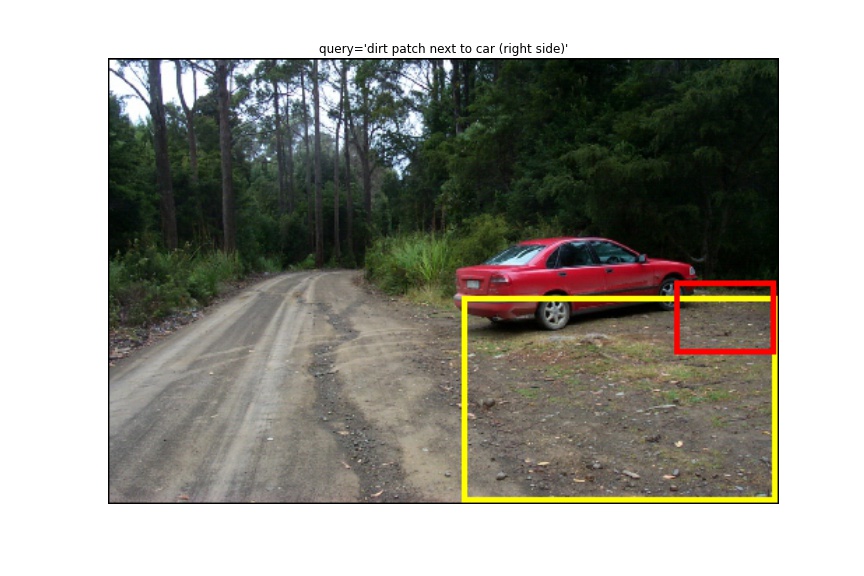}

\end{tabular}
\end{center}
\caption{Failure cases ($\mathrm{IoU} < 0.5$) on ReferIt with EdgeBox. Ground truth in yellow and incorrectly retrieved box in red. Some failures cases are caused by ambiguity of the query and some due to wrong annotations in the dataset. }
\label{fig:referit_failure}
\end{figure*}

Table \ref{tab:results_referit_edgebox} shows the result of the second scenario on 100 EdgeBox proposals, where ``R@1'' is the recall of the highest scoring box (the percentage of the highest scoring box being correct), and ``R@10'' is the percentage of at least one of the 10 highest scoring proposals being correct. We also report ``Oracle'' (or equivalently ``R@100''), the percentage of at least one of all 100 proposals being correct, as an upper-bound of all object retrieval systems in this scenario. It can be seen that results in Table \ref{tab:results_referit_edgebox} follow the same trend as in Table \ref{tab:results_referit_gt}, with our full SCRC model achieving the highest recall. Figure \ref{fig:referit_success} shows examples of correctly retrieved objects at top-1 using 100 EdgeBox proposals, where the highest scoring candidate region from our SCRC model overlaps with ground truth annotation by at least $50\%$ IoU, and Figure \ref{fig:referit_failure} shows some failure cases, where retrieved top-1 candidate region fails to match ground truth.

By comparing ``SCRC (w/o context, spatial)'' and ``SCRC (w/o context, spatial, transfer)'' in Table \ref{tab:results_referit_gt} and Table \ref{tab:results_referit_edgebox}, it can also be seen that the pretraining and adaptation procedure described in Section \ref{sec:method} outperforms directly training on retrieval dataset, showing that pretraining allows the model to transfer useful visual-linguistic knowledge from image captioning dataset.

Also, our SCRC model outperforms the bag-of-words CAFFE-7K model and LRCN model significantly. Compared with our model, CAFFE-7K method suffers from information loss by first projecting image region to category names and limited vocabulary drawn from predefined object category names, and is not end-to-end trainable. Although LRCN model trained on MSCOCO for image captioning task is effective for text-based image retrieval as shown in \cite{donahue2015long}, directly running it as an object retriever on a set of candidate boxes results in inferior performance. This is because object retrieval and image retrieval are different domains, and LRCN model as a object retriever does not encode spatial configuration or global context.

\begin{table}[t]
\begin{center}
\small
\begin{tabular}{|l|c|c|}
\hline
\textbf{Method} & R@1 & R@10 \\
\hline
CAFFE-7K \cite{guadarrama2014open}              &  10.38\%           &  26.20\%           \\
LRCN \cite{donahue2015long}                     & ~~8.59\%           &  31.86\%           \\
SCRC (w/o context, spatial, transfer)           &  14.53\%           &  40.72\%           \\
SCRC (w/o context, spatial)                     &  15.78\%           &  42.54\%           \\
SCRC (w/o context)                              &  17.68\%           &  44.77\%           \\
SCRC                                            &  \textbf{17.93\%}  &  \textbf{45.27\%}  \\
\hline
Oracle                                          &  59.38\%           &  59.38\%           \\
\hline
\end{tabular}
\end{center}
\caption{Performance of our method compared with baselines on 100 EdgeBox proposals in ReferIt dataset. See \secref{sec:referit:results} for details.}
\label{tab:results_referit_edgebox}
\end{table}

\begin{figure*}[t]
\begin{center}
\begin{tabular}{c@{}c@{}c@{}c}
\small{a scene with three people} &
\small{query='\textit{man far right}'} &
\small{query='\textit{left guy}'} &
\small{query='\textit{cyclist}'} \\
\includegraphics[trim = 95mm 25mm 95mm 20mm , clip=true,width=0.18\linewidth,height=\textheight,keepaspectratio]{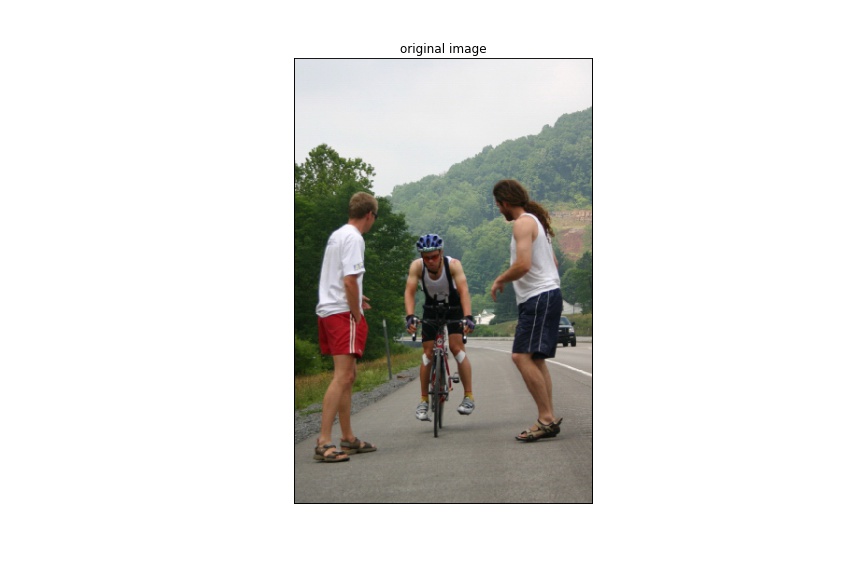} &
\includegraphics[trim = 95mm 25mm 95mm 20mm , clip=true,width=0.18\linewidth,height=\textheight,keepaspectratio]{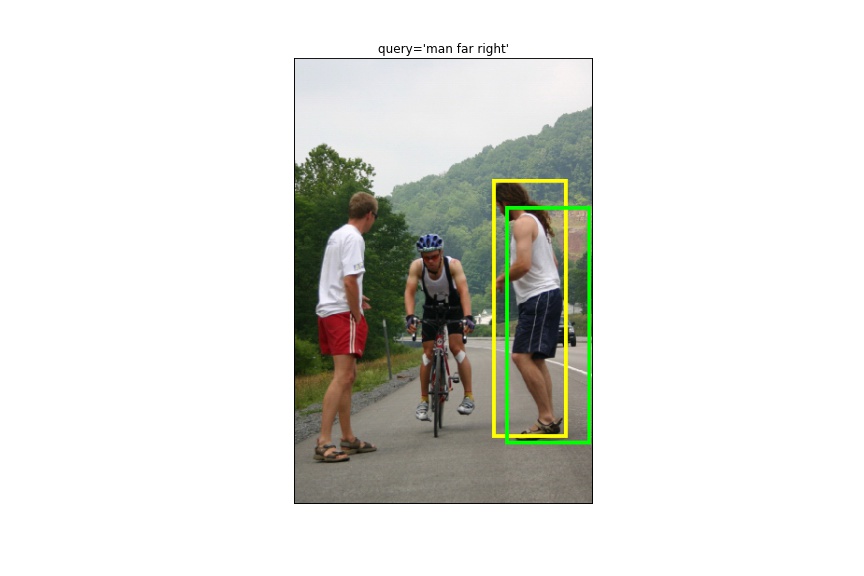} &
\includegraphics[trim = 95mm 25mm 95mm 20mm , clip=true,width=0.18\linewidth,height=\textheight,keepaspectratio]{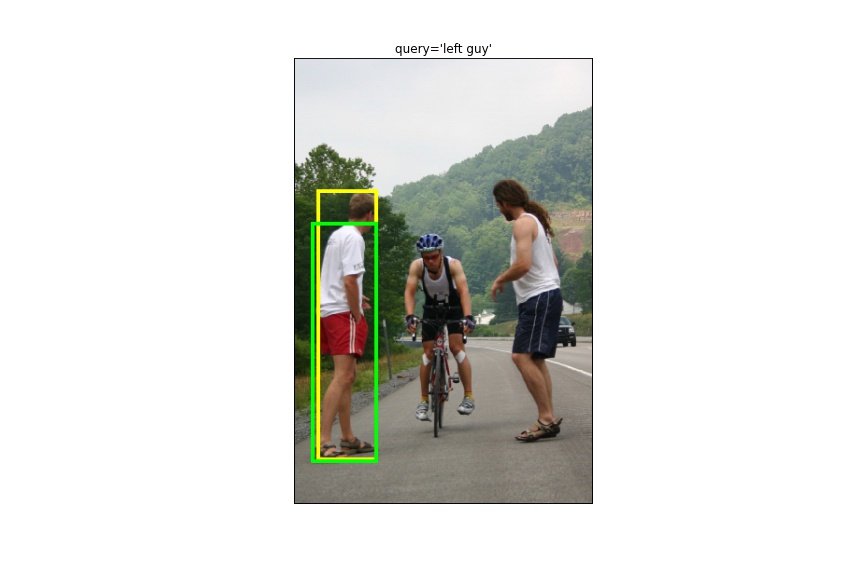} &
\includegraphics[trim = 95mm 25mm 95mm 20mm , clip=true,width=0.18\linewidth,height=\textheight,keepaspectratio]{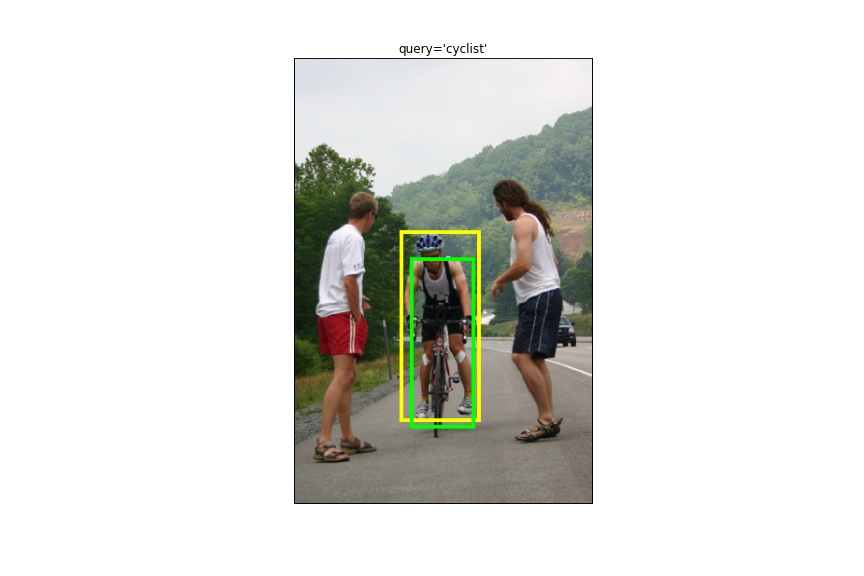} \\
\end{tabular}
\end{center}
\caption{An example image in ReferIt dataset where objects are described based on other objects in the scene. When referring to one of the three ``people'' in the image, expressions based on both the object and the context are used to make the description discriminative. Our model can handle such object descriptions in context by incorporating these information into the recurrent neural network. In the images above, yellow boxes are ground truth and green boxes are correctly retrieved results by our model using highest scoring candidate from 100 EdgeBox proposals. }
\label{fig:referit_context_eg}
\end{figure*}

\mysubsubsection{Retrieval on object descriptions in context}
In reality, people usually describe an object based on both the object itself and other objects plus the whole scene as context. To distinguish a specific object from others in a scene, especially when there are multiple objects of the same category, a description needs to contain not only the category name, but also other discriminative information such as locations or attributes.

\begin{figure*}
\begin{center}
\begin{tabular}{c@{}c@{}c}

\small{query='\textit{far right person}'} & \multicolumn{1}{m{5.0cm}}{\small{query='\textit{lady very back with white shirts on, next to man in hat}'}} & \small{query='\textit{lady in black shirt}'} \\
\includegraphics[trim = 40mm 25mm 40mm 20mm , clip=true,width=0.3\textwidth,height=\textheight,keepaspectratio]{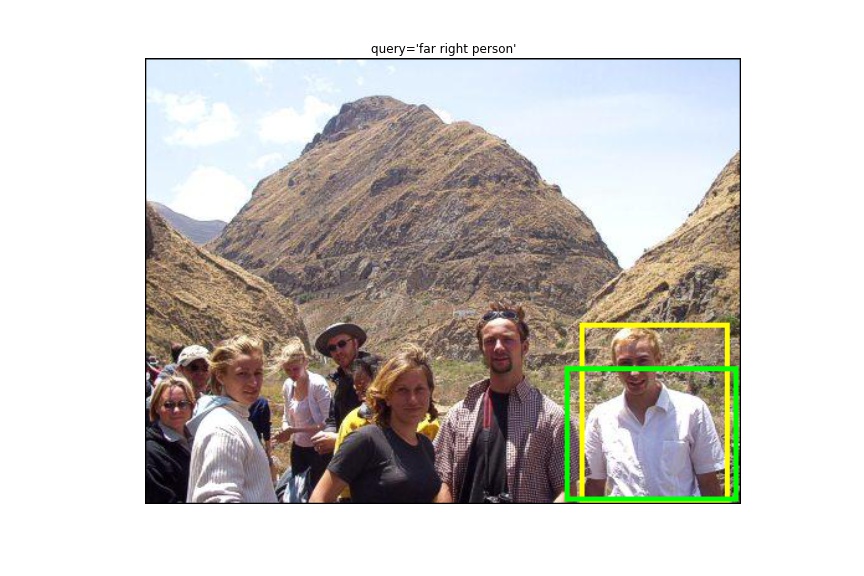} & 
\includegraphics[trim = 40mm 25mm 40mm 20mm , clip=true,width=0.3\textwidth,height=\textheight,keepaspectratio]{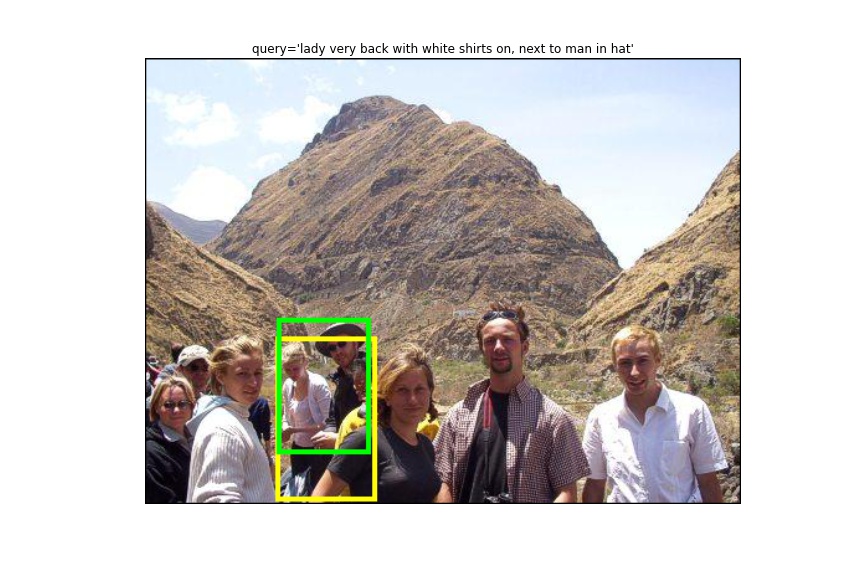} &
\includegraphics[trim = 40mm 25mm 40mm 20mm , clip=true,width=0.3\textwidth,height=\textheight,keepaspectratio]{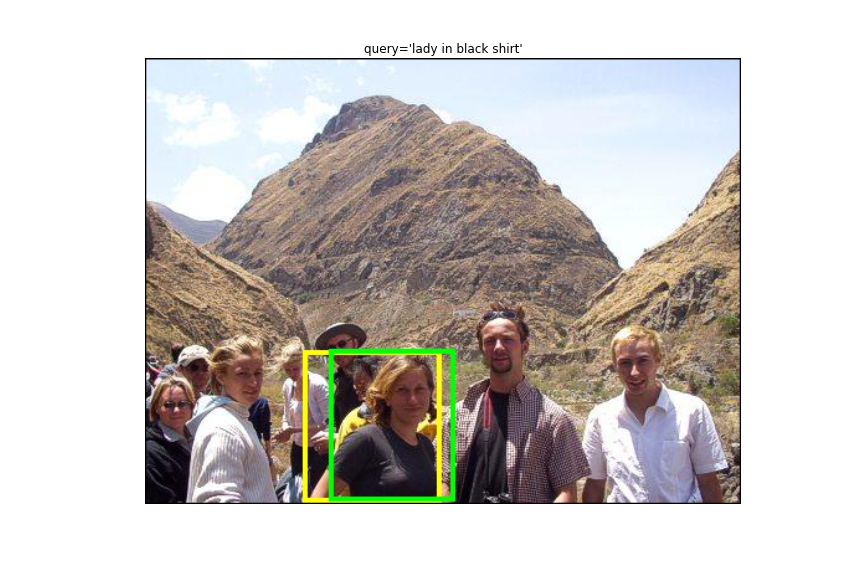} \\

\small{query='\textit{bottom left window}'} & \small{query='\textit{fenced window left of center door}'} & \small{query='\textit{window upper right}'} \\
\includegraphics[trim = 40mm 25mm 40mm 20mm , clip=true,width=0.3\textwidth,height=\textheight,keepaspectratio]{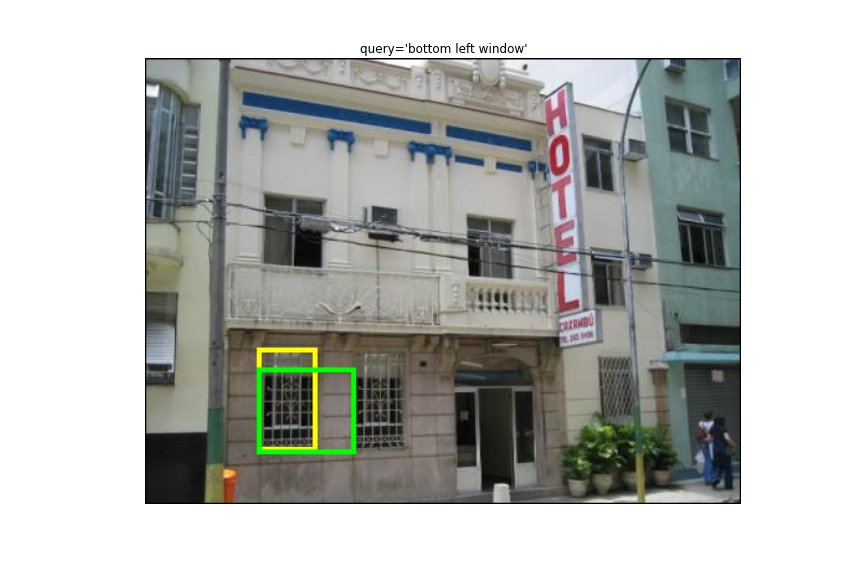} & 
\includegraphics[trim = 40mm 25mm 40mm 20mm , clip=true,width=0.3\textwidth,height=\textheight,keepaspectratio]{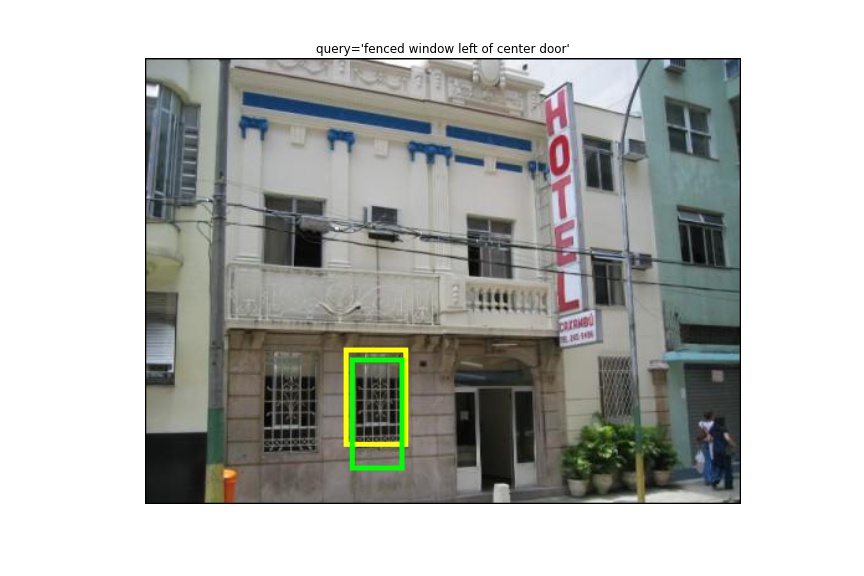} &
\includegraphics[trim = 40mm 25mm 40mm 20mm , clip=true,width=0.3\textwidth,height=\textheight,keepaspectratio]{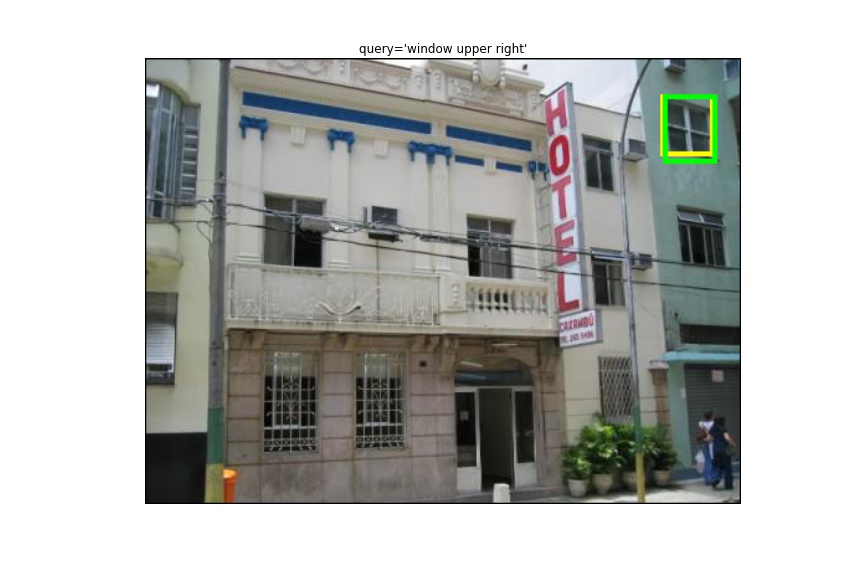} \\

\small{query='\textit{2 people on left}'} & \small{query='\textit{dude center with backpack blue}'} & \multicolumn{1}{m{5.0cm}}{\small{query='\textit{guy with the tan pants and backpack}'}} \\
\includegraphics[trim = 40mm 25mm 40mm 20mm , clip=true,width=0.3\textwidth,height=\textheight,keepaspectratio]{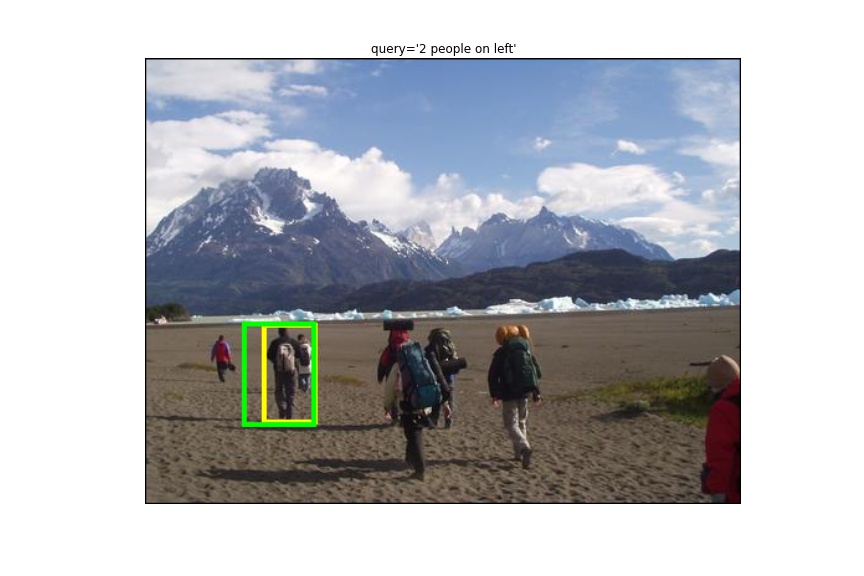} &
\includegraphics[trim = 40mm 25mm 40mm 20mm , clip=true,width=0.3\textwidth,height=\textheight,keepaspectratio]{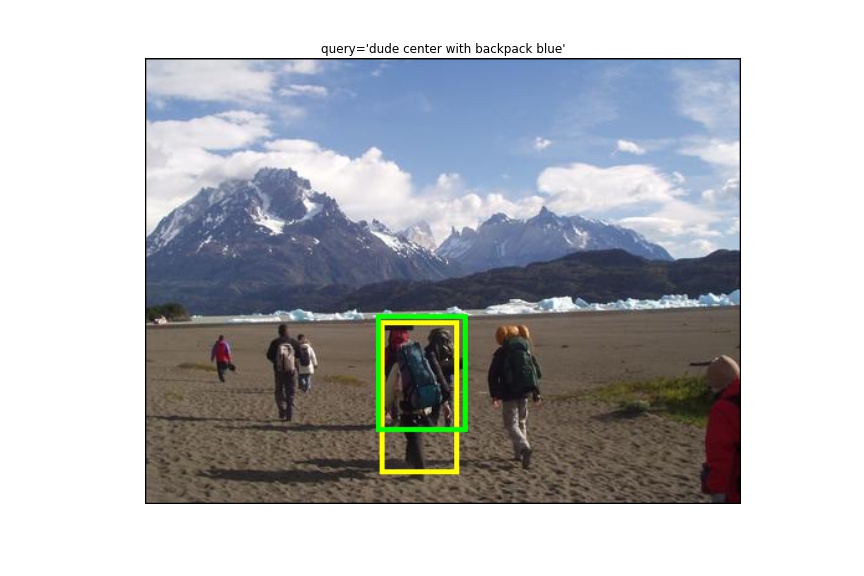} &
\includegraphics[trim = 40mm 25mm 40mm 20mm , clip=true,width=0.3\textwidth,height=\textheight,keepaspectratio]{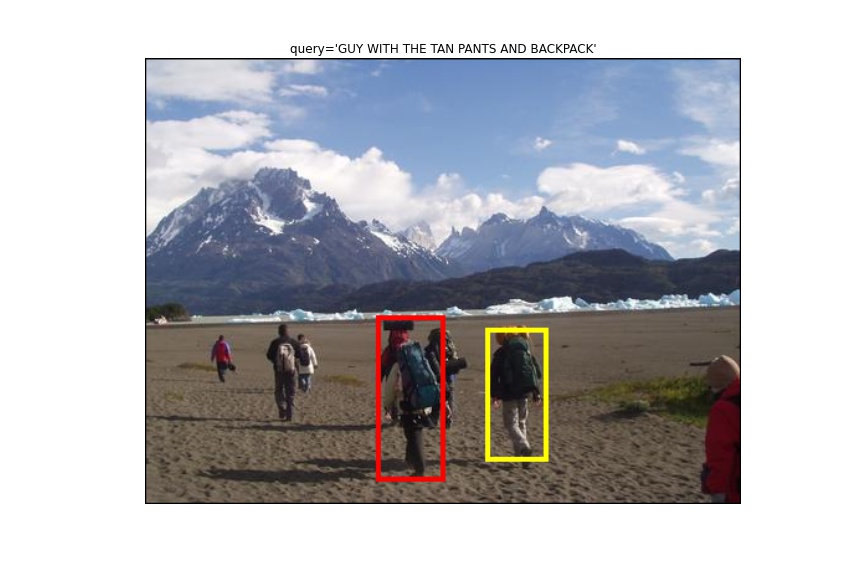}\\

\small{query='\textit{chair left}'} & \small{query='\textit{nice plush chair}'} & \small{query='\textit{lamp}'} \\
\includegraphics[trim = 40mm 25mm 40mm 20mm , clip=true,width=0.3\textwidth,height=\textheight,keepaspectratio]{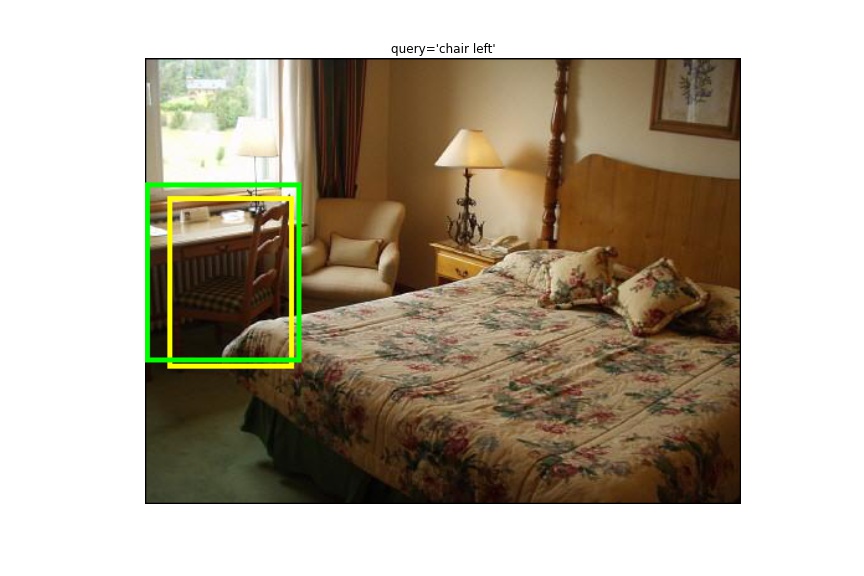} & 
\includegraphics[trim = 40mm 25mm 40mm 20mm , clip=true,width=0.3\textwidth,height=\textheight,keepaspectratio]{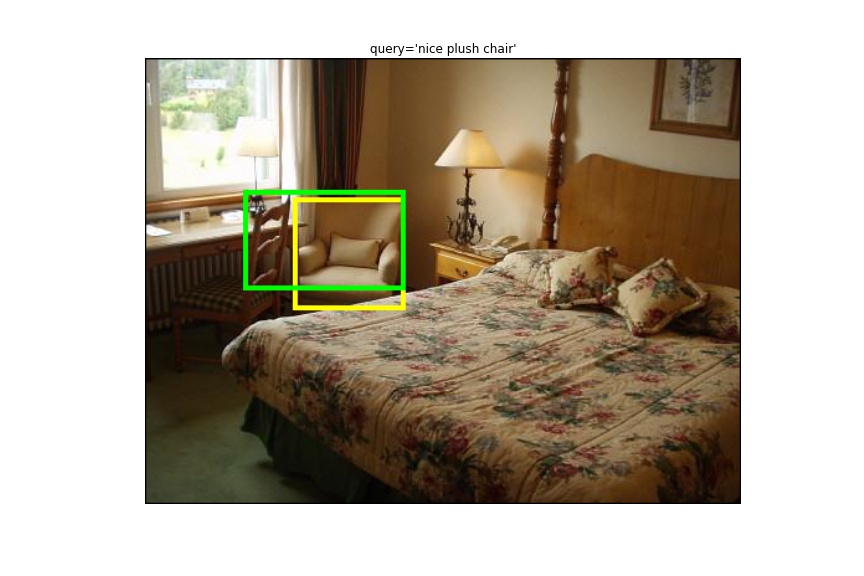} &
\includegraphics[trim = 40mm 25mm 40mm 20mm , clip=true,width=0.3\textwidth,height=\textheight,keepaspectratio]{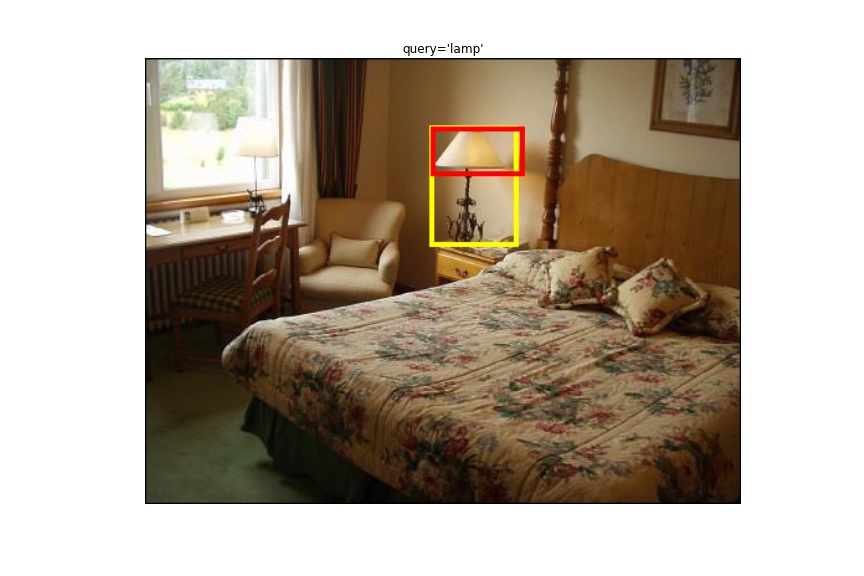} \\

\small{query='\textit{picture 2nd from left}'} & \small{query='\textit{third picture from left}'} & \small{query='\textit{picture second from right}'} \\
\includegraphics[trim = 40mm 25mm 40mm 20mm , clip=true,width=0.3\textwidth,height=\textheight,keepaspectratio]{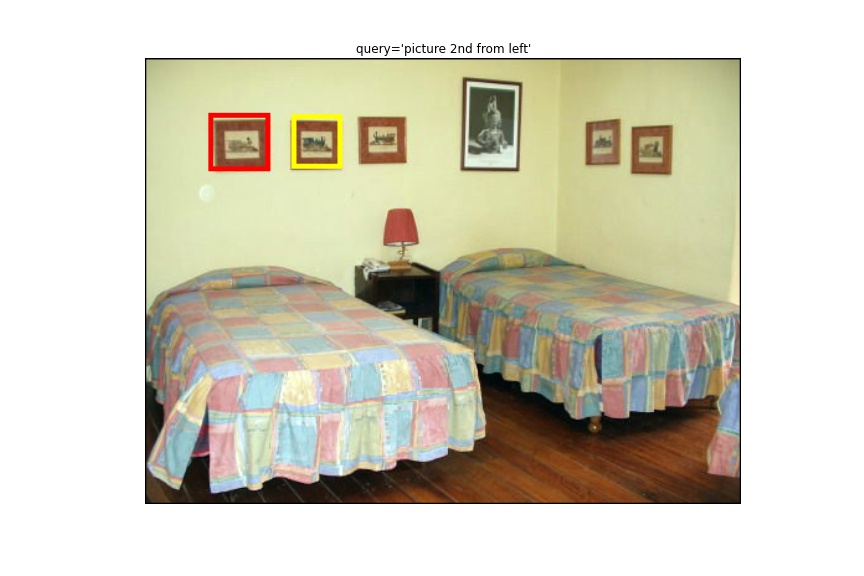} &
\includegraphics[trim = 40mm 25mm 40mm 20mm , clip=true,width=0.3\textwidth,height=\textheight,keepaspectratio]{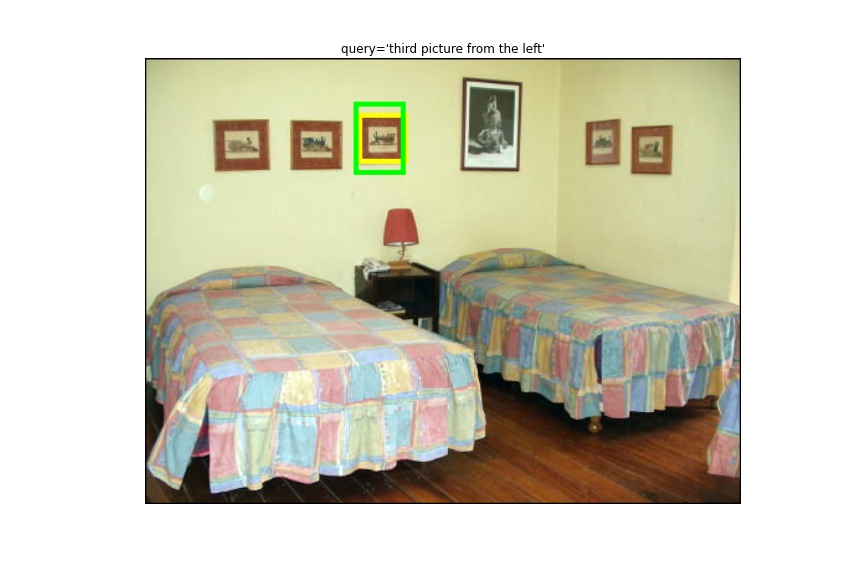} &
\includegraphics[trim = 40mm 25mm 40mm 20mm , clip=true,width=0.3\textwidth,height=\textheight,keepaspectratio]{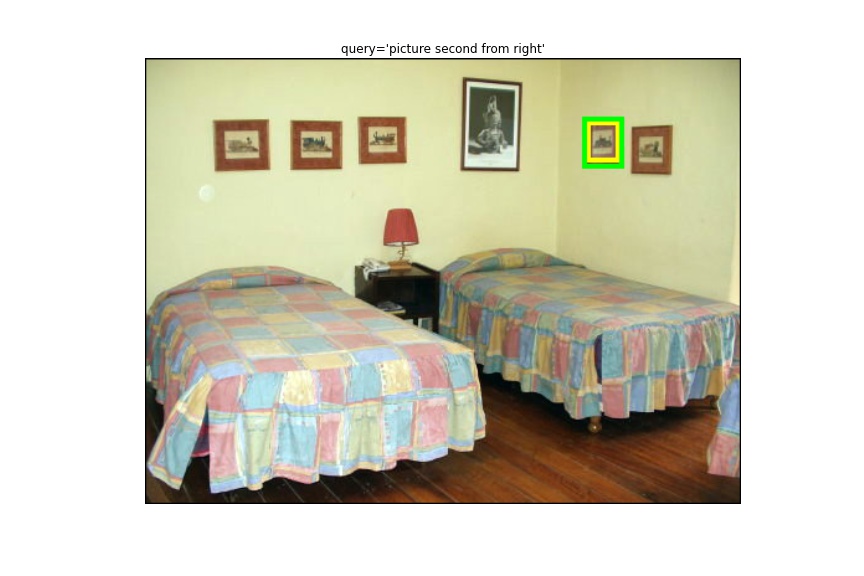} \\

\end{tabular}
\end{center}
\caption{Examples on multiple objects in the same image in ReferIt, showing the highest scoring candidate box (correct in green, incorrect in red) from 100 EdgeBox proposals and ground truth (yellow). Our model retrieves the objects by taking their local descriptors, spatial configurations and scene-level contextual information into account. }
\label{fig:referit_context_multiple}
\end{figure*}

Figure \ref{fig:referit_context_eg} shows an example of this, where one cannot refer to a person simply using category name ``person'' since there are three people in the scene, but needs to use a description based on the environment as query. Our SCRC model can handle such context-based descriptions by incorporating spatial configurations and scene-level context into the recurrent network. Figure \ref{fig:referit_context_multiple} shows some retrieval examples on multiple objects within the same image on ReferIt \cite{kazemzadeh2014referitgame} dataset, where objects are described in context.

\mysubsubsection{Object vs. stuff}
The ReferIt dataset contains annotations on both ``object'' regions and ``stuff'' regions. In computer vision, the term \textbf{object} is usually used to refer to entities with closed boundary and well-defined shape, such as ``car'', ``person'' and ``laptop''. On the other hand, \textbf{stuff} is used for entities without a regular shape, such as ``grass'', ``road'' and ``sky''.

\begin{figure*}
\begin{center}
\begin{tabular}{c@{}c@{}c}

\small{query='\textit{river}'} & \small{query='\textit{sky between flags}'} & \small{query='\textit{grass upper right}'} \\
\includegraphics[trim = 40mm 25mm 40mm 20mm , clip=true,width=0.3\textwidth,height=\textheight,keepaspectratio]{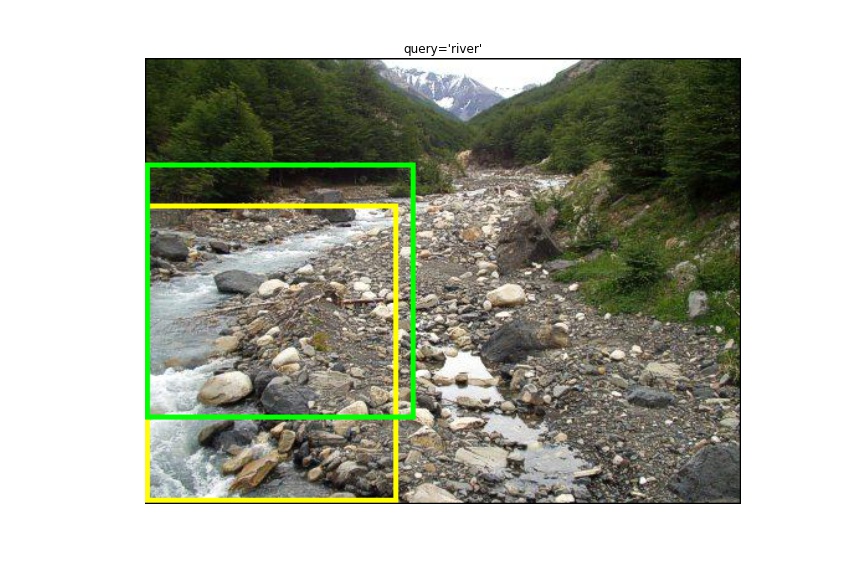} &
\includegraphics[trim = 40mm 25mm 40mm 20mm , clip=true,width=0.3\textwidth,height=\textheight,keepaspectratio]{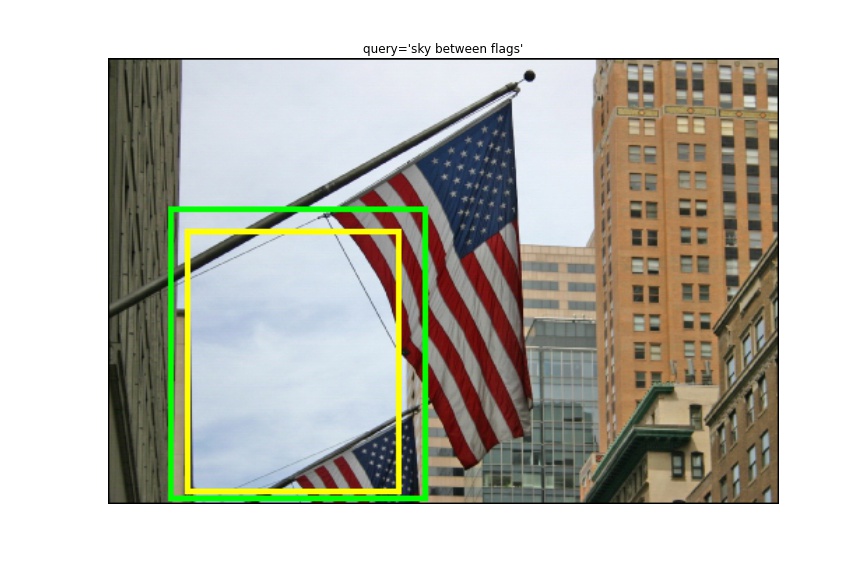} & 
\includegraphics[trim = 40mm 25mm 40mm 20mm , clip=true,width=0.3\textwidth,height=\textheight,keepaspectratio]{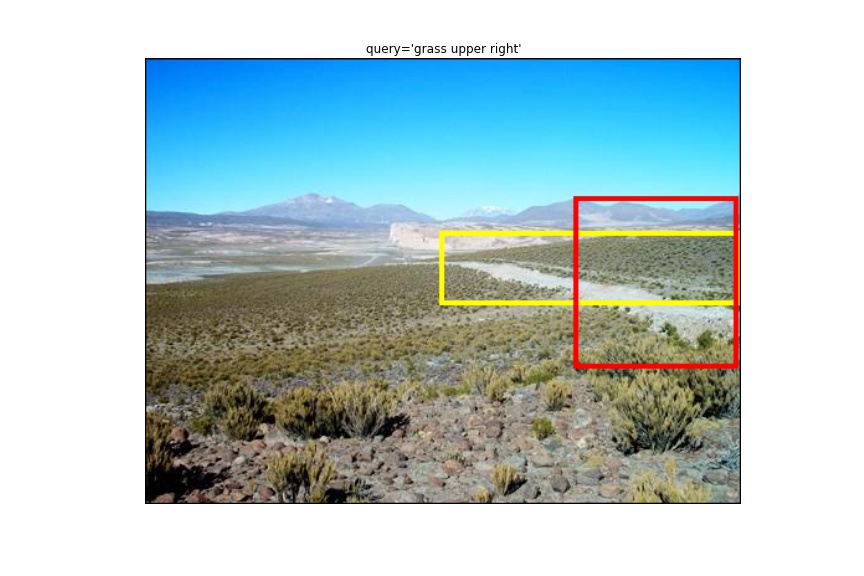} \\

\small{query='\textit{city}'} & \small{query='\textit{big gray roof at bottom}'} & \multicolumn{1}{m{5.0cm}}{\small{query='\textit{rocky wall directly to the right of people}'}} \\
\includegraphics[trim = 40mm 25mm 40mm 20mm , clip=true,width=0.3\textwidth,height=\textheight,keepaspectratio]{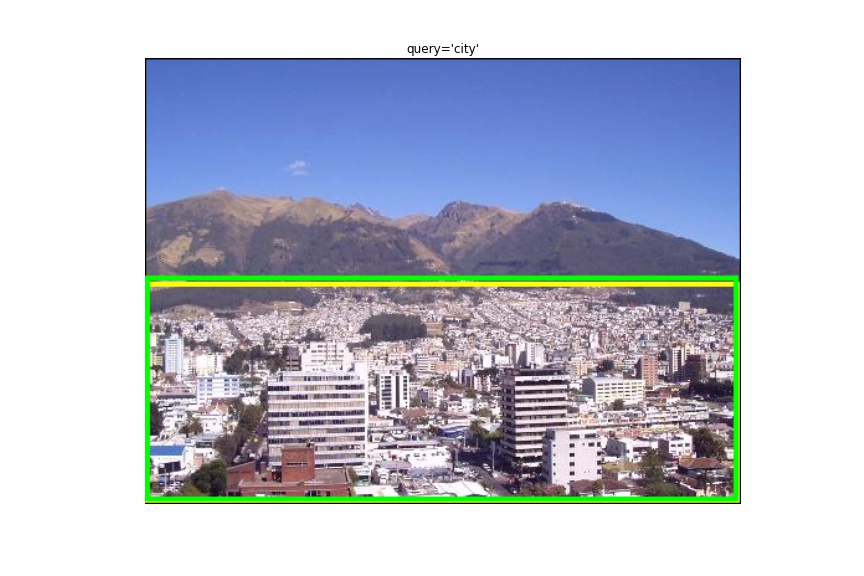} & 
\includegraphics[trim = 40mm 25mm 40mm 20mm , clip=true,width=0.3\textwidth,height=\textheight,keepaspectratio]{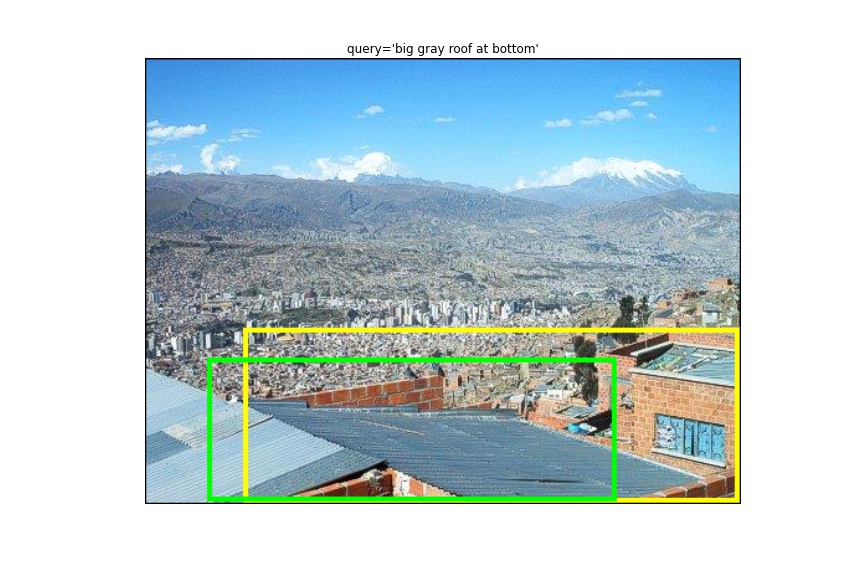} &
\includegraphics[trim = 40mm 25mm 40mm 20mm , clip=true,width=0.3\textwidth,height=\textheight,keepaspectratio]{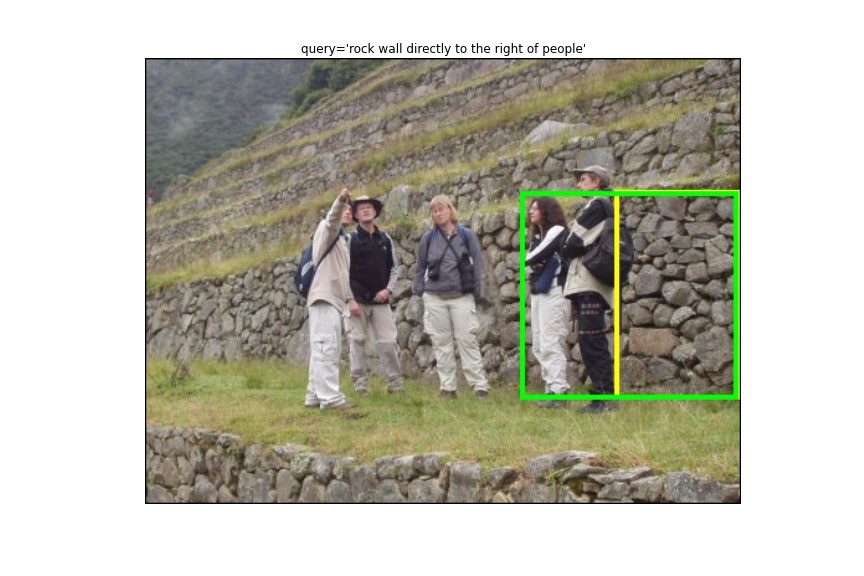} \\

\small{query='\textit{road in front of biking dude}'} & \small{query='\textit{hill}'} & \multicolumn{1}{m{5.0cm}}{\small{query='\textit{left white sky just above dark mountain}'}} \\
\includegraphics[trim = 40mm 25mm 40mm 20mm , clip=true,width=0.3\textwidth,height=\textheight,keepaspectratio]{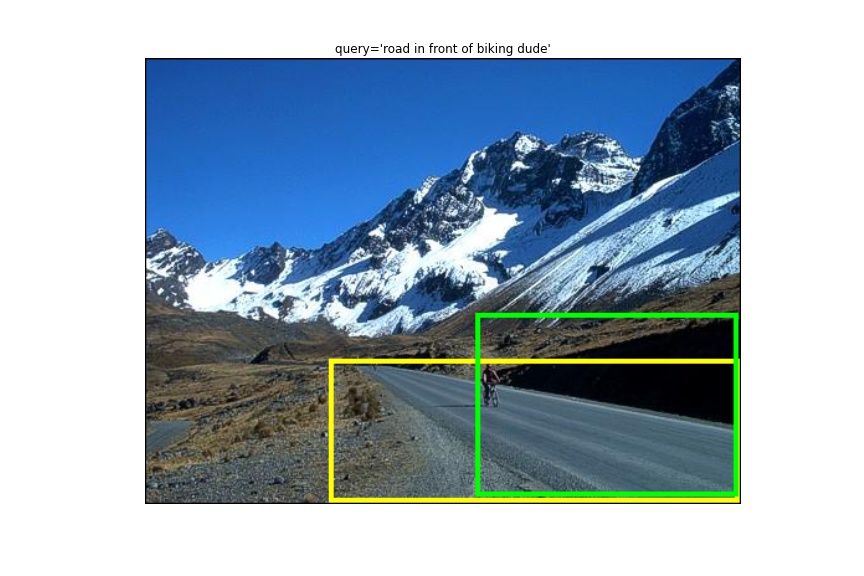} & 
\includegraphics[trim = 40mm 25mm 40mm 20mm , clip=true,width=0.3\textwidth,height=\textheight,keepaspectratio]{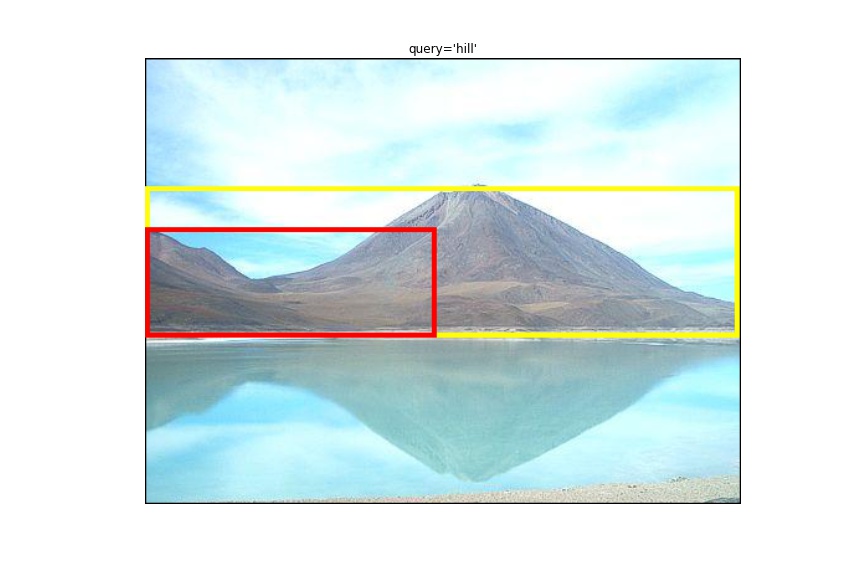} &
\includegraphics[trim = 40mm 25mm 40mm 20mm , clip=true,width=0.3\textwidth,height=\textheight,keepaspectratio]{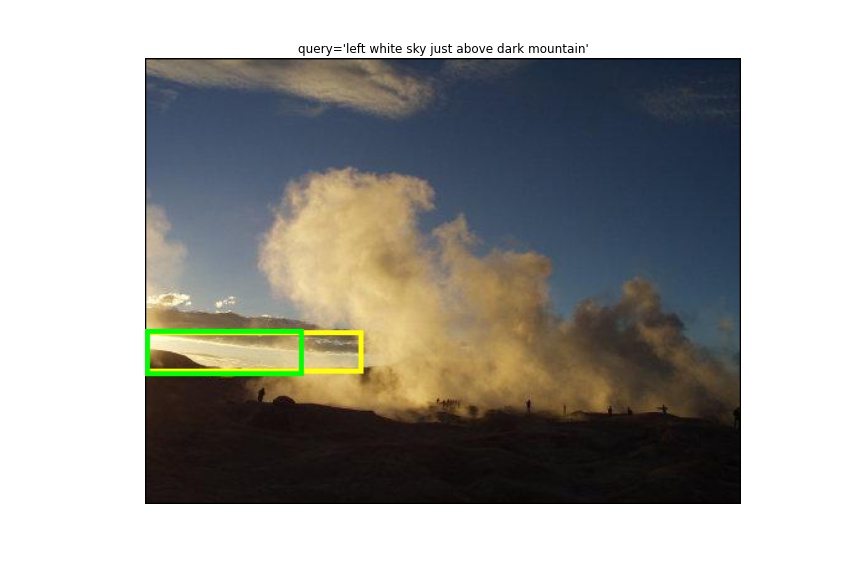} \\

\multicolumn{1}{m{5.0cm}}{\small{query='\textit{grass to the right of the right front tire}'}} & \multicolumn{1}{m{5.0cm}}{\small{query='\textit{lake underneath the mountain on the left}'}} & \multicolumn{1}{m{5.0cm}}{\small{query='\textit{mountain left side above peoples head}'}} \\
\includegraphics[trim = 40mm 25mm 40mm 20mm , clip=true,width=0.3\textwidth,height=\textheight,keepaspectratio]{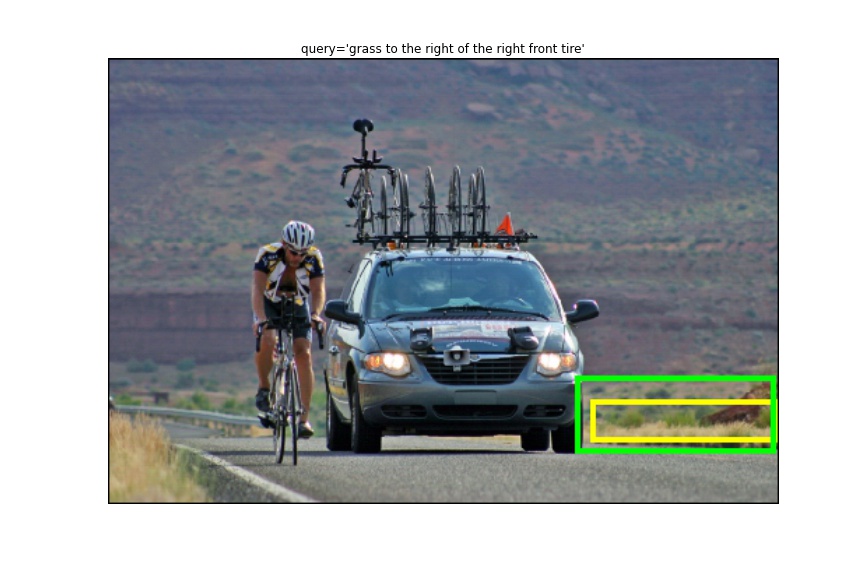} & 
\includegraphics[trim = 40mm 25mm 40mm 20mm , clip=true,width=0.3\textwidth,height=\textheight,keepaspectratio]{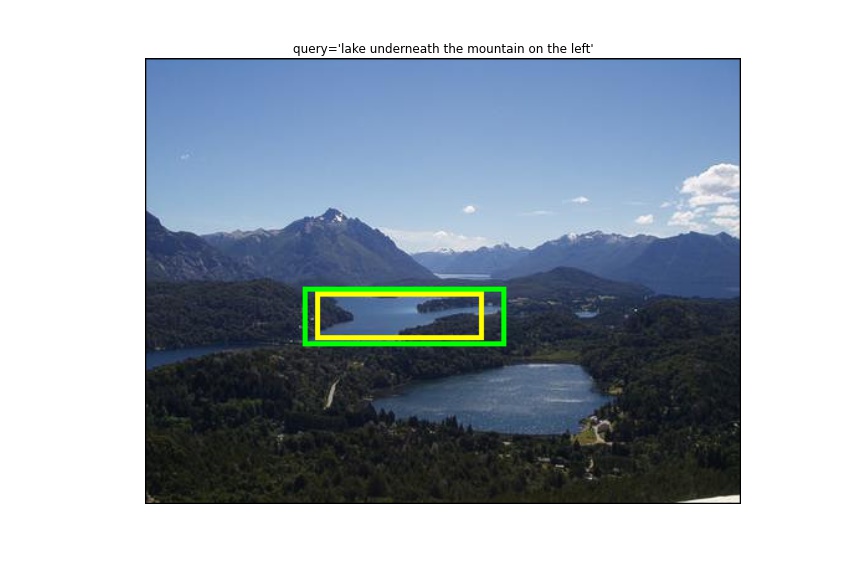} &
\includegraphics[trim = 40mm 25mm 40mm 20mm , clip=true,width=0.3\textwidth,height=\textheight,keepaspectratio]{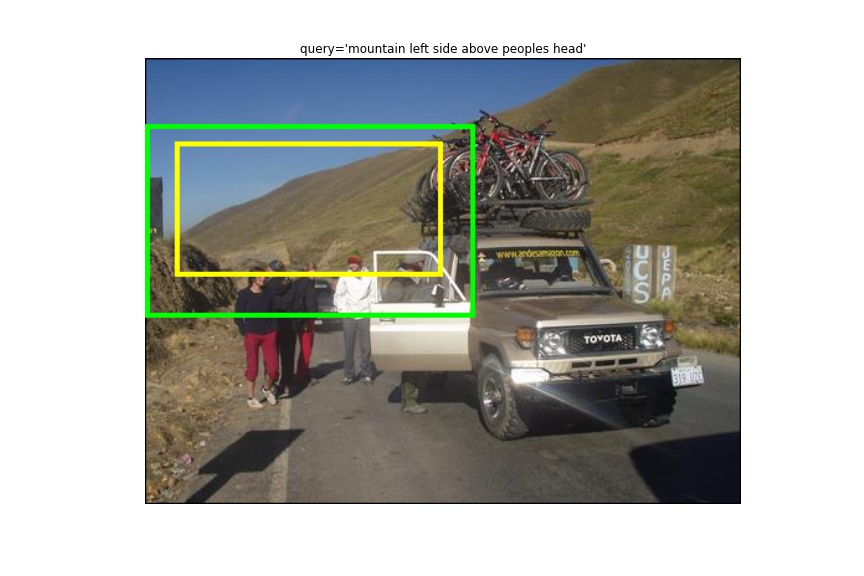} \\

\small{query='\textit{the smoke}'} & \small{query='\textit{water in the center bottom}'} & \small{query='\textit{sand}'} \\
\includegraphics[trim = 40mm 25mm 40mm 20mm , clip=true,width=0.3\textwidth,height=\textheight,keepaspectratio]{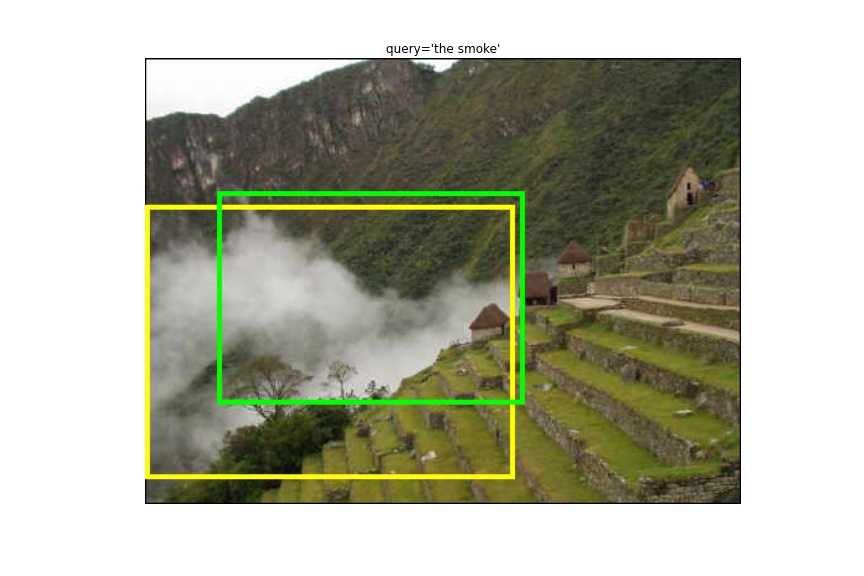} & 
\includegraphics[trim = 40mm 25mm 40mm 20mm , clip=true,width=0.3\textwidth,height=\textheight,keepaspectratio]{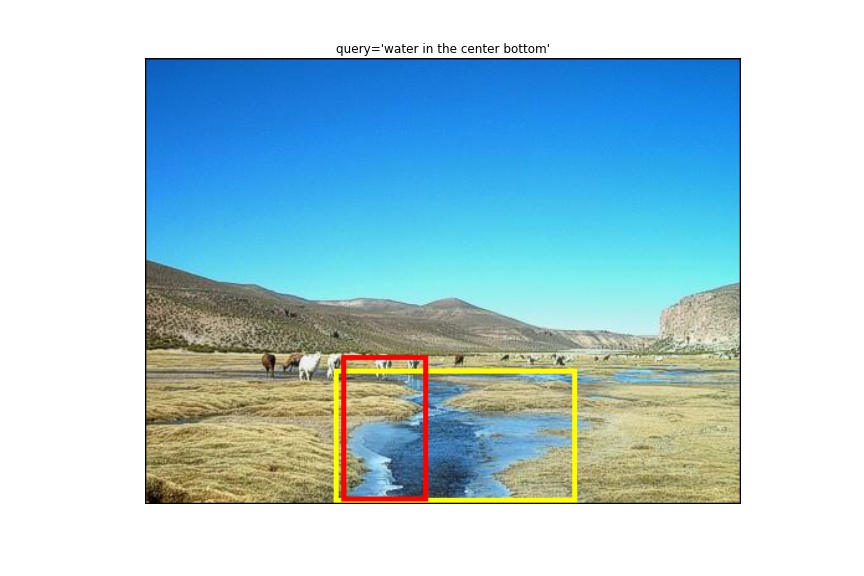} &
\includegraphics[trim = 40mm 25mm 40mm 20mm , clip=true,width=0.3\textwidth,height=\textheight,keepaspectratio]{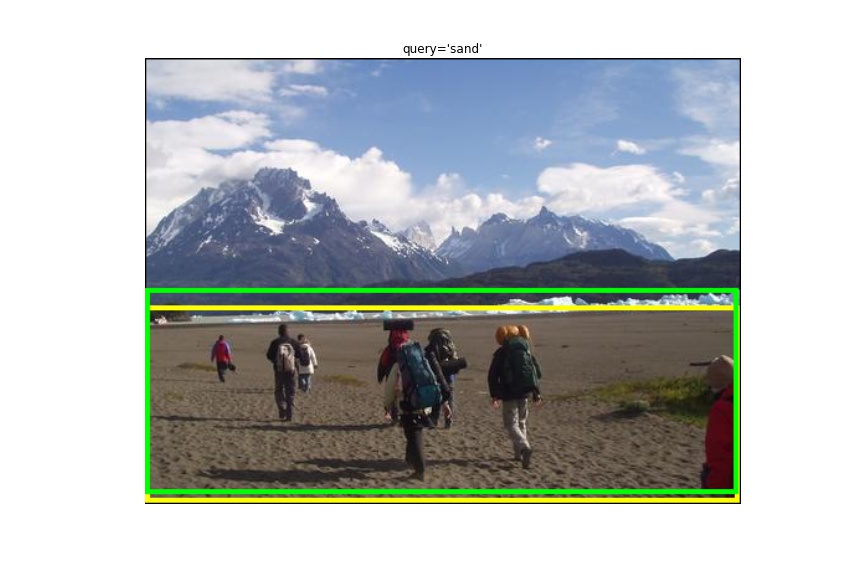} \\

\end{tabular}
\end{center}
\caption{Examples on ``stuff'' regions in ReferIt, showing the highest scoring candidate box (correct in green, incorrect in red) from 100 EdgeBox proposals and ground truth (yellow).}
\label{fig:referit_stuff}
\end{figure*}

Given an input image and a natural language query, our SCRC model is not only capable of retrieving ``object'' regions, but can also be applied to ``stuff'' regions. Figure \ref{fig:referit_stuff} shows some examples of stuff retrieval on ReferIt dataset.

\begin{figure*}
\begin{center}
\begin{tabular}{c@{}c@{}c}

{\small{generated description='\textit{bed on left}'}} &
{\small{generated description='\textit{yellow car}'}} &
{\small{generated description='\textit{horse}'}} \\
\includegraphics[trim = 40mm 25mm 40mm 20mm , clip=true,width=0.3\textwidth,height=\textheight,keepaspectratio]{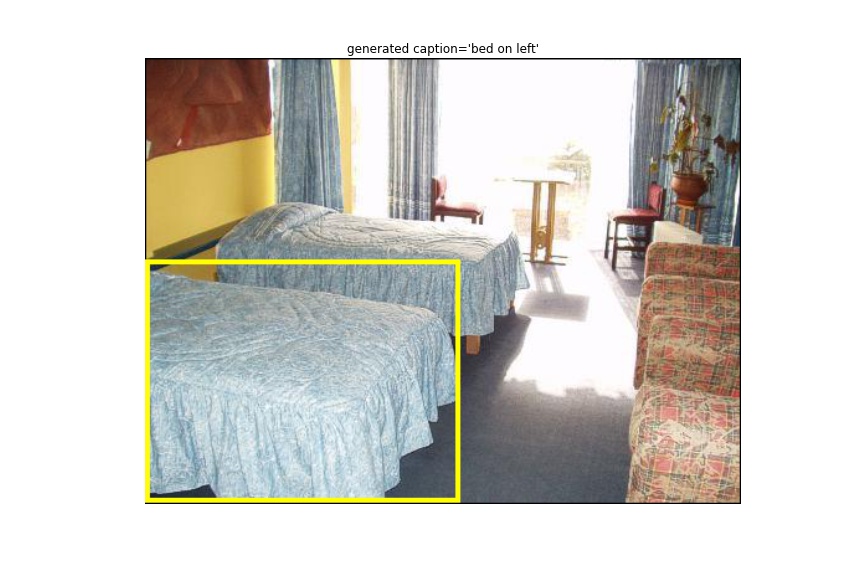} &
\includegraphics[trim = 40mm 25mm 40mm 20mm , clip=true,width=0.3\textwidth,height=\textheight,keepaspectratio]{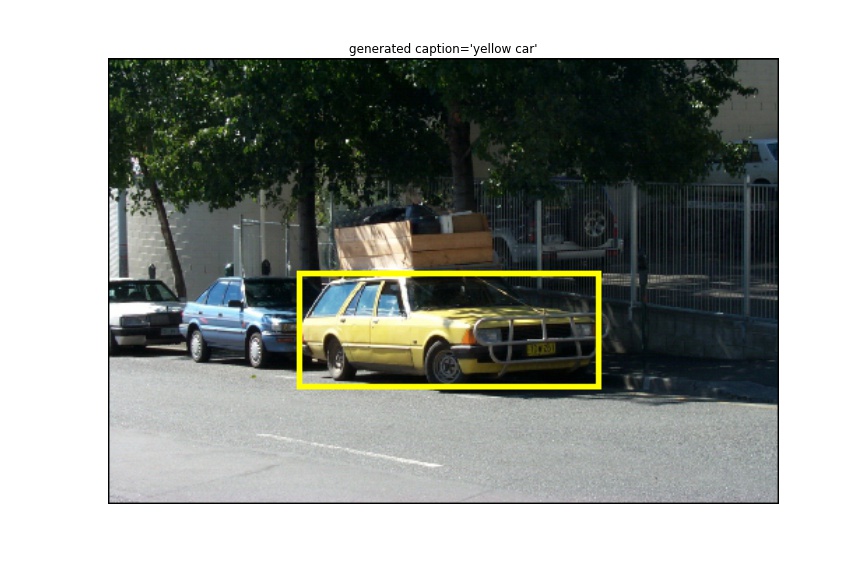} & 
\includegraphics[trim = 40mm 25mm 40mm 20mm , clip=true,width=0.3\textwidth,height=\textheight,keepaspectratio]{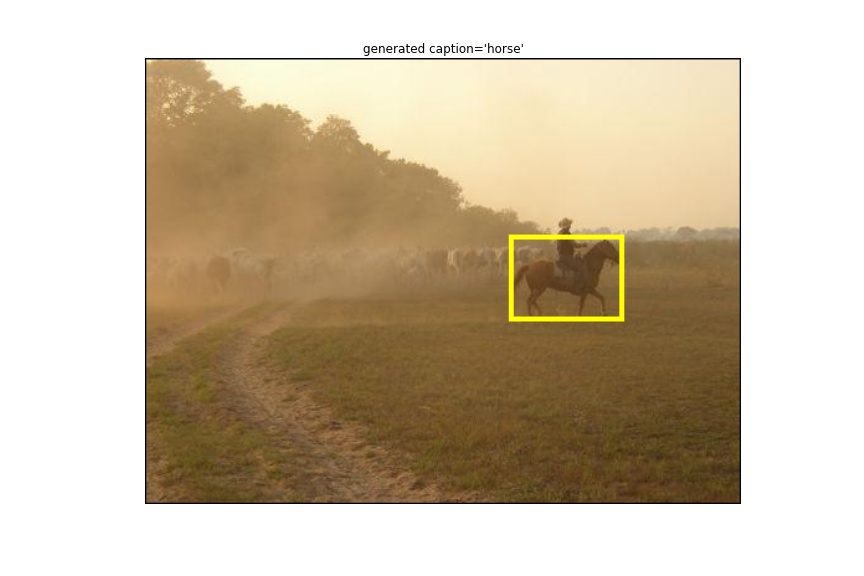} \\

{\small{generated description='\textit{man in blue shirt}'}} &
{\small{generated description='\textit{red backpack}'}} &
{\small{generated description='\textit{snow}'}} \\
\includegraphics[trim = 40mm 25mm 40mm 20mm , clip=true,width=0.3\textwidth,height=\textheight,keepaspectratio]{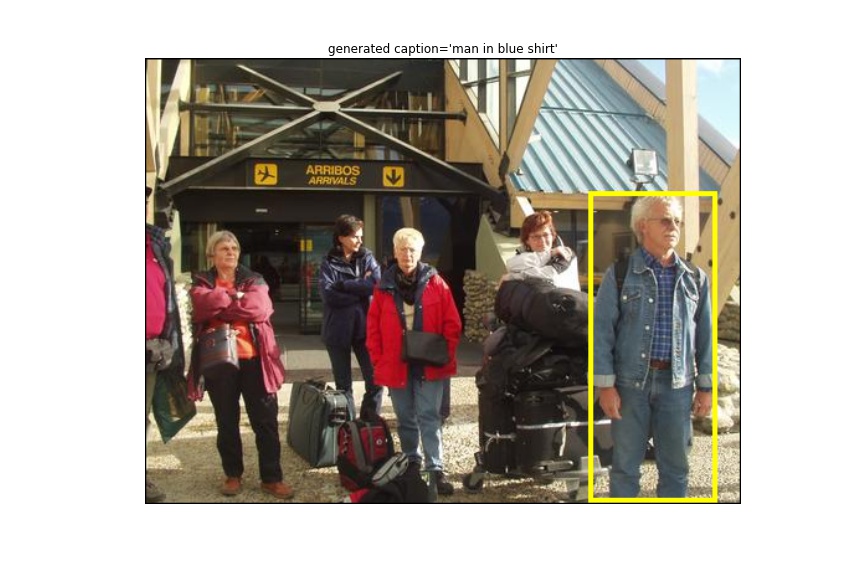} & 
\includegraphics[trim = 40mm 25mm 40mm 20mm , clip=true,width=0.3\textwidth,height=\textheight,keepaspectratio]{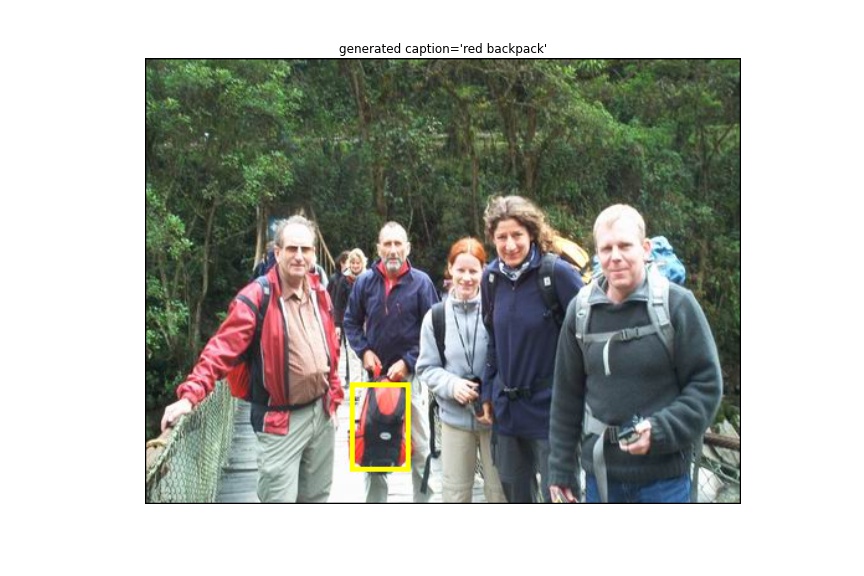} &
\includegraphics[trim = 40mm 25mm 40mm 20mm , clip=true,width=0.3\textwidth,height=\textheight,keepaspectratio]{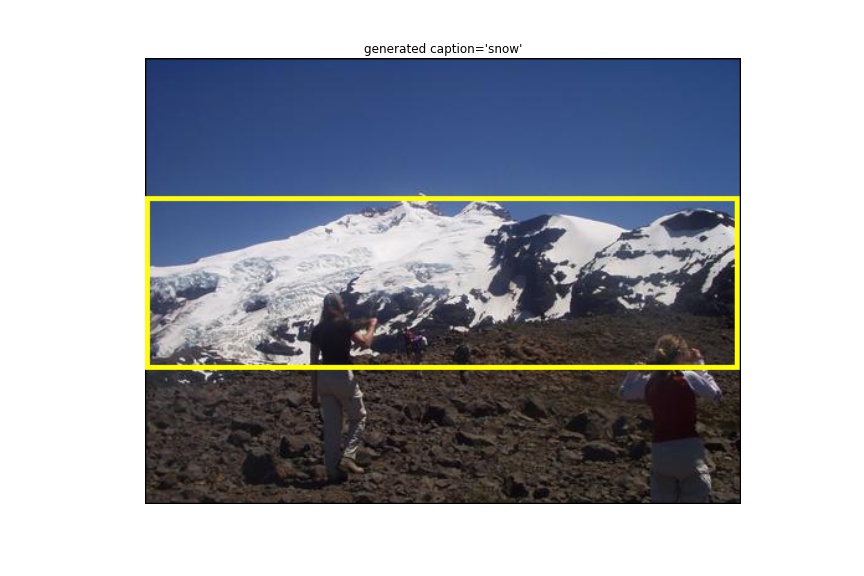} \\

{\small{generated description='\textit{tree on the left}'}} &
{\small{generated description='\textit{door}'}} &
{\small{generated description='\textit{clouds}'}} \\
\includegraphics[trim = 40mm 25mm 40mm 20mm , clip=true,width=0.3\textwidth,height=\textheight,keepaspectratio]{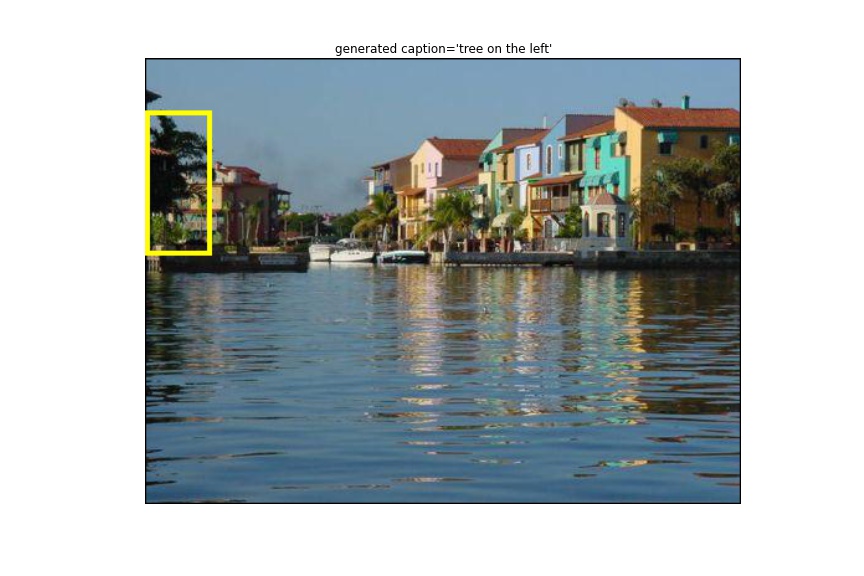} &
\includegraphics[trim = 40mm 25mm 40mm 20mm , clip=true,width=0.3\textwidth,height=\textheight,keepaspectratio]{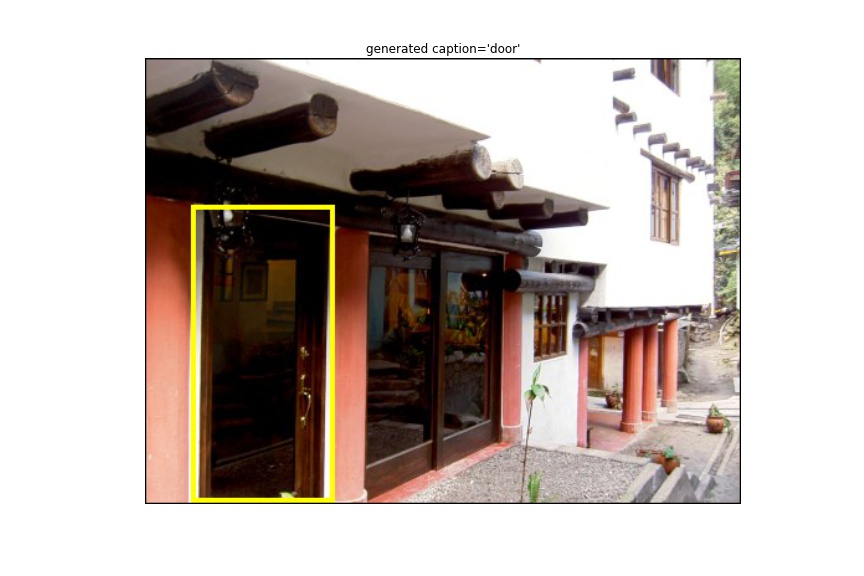} & 
\includegraphics[trim = 40mm 25mm 40mm 20mm , clip=true,width=0.3\textwidth,height=\textheight,keepaspectratio]{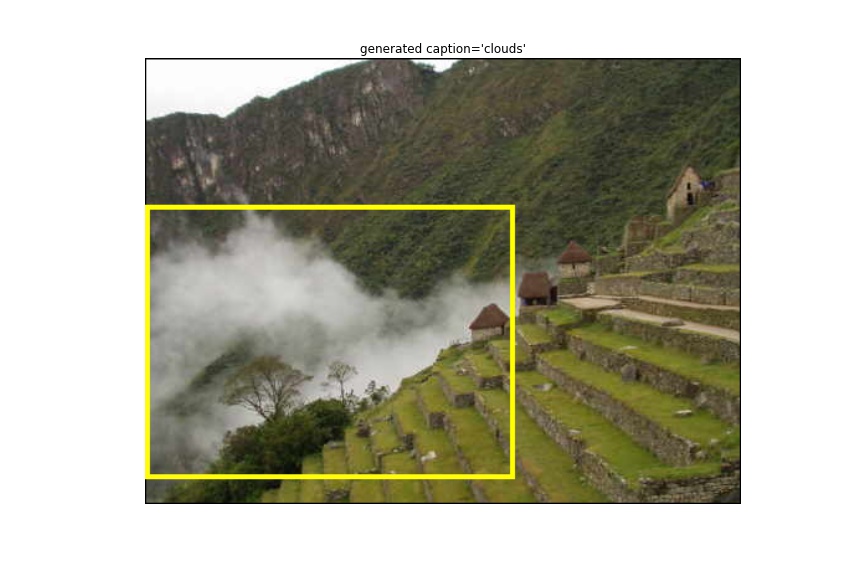} \\

\multicolumn{1}{m{5.0cm}}{\small{generated description='\textit{hat on the woman in red}'}} &
\multicolumn{1}{m{5.0cm}}{\small{generated description='\textit{desk in front of kid with red shirt}'}} &
\multicolumn{1}{m{5.0cm}}{\small{generated description='\textit{plant in front of pink vase}'}} \\
\includegraphics[trim = 40mm 25mm 40mm 20mm , clip=true,width=0.3\textwidth,height=\textheight,keepaspectratio]{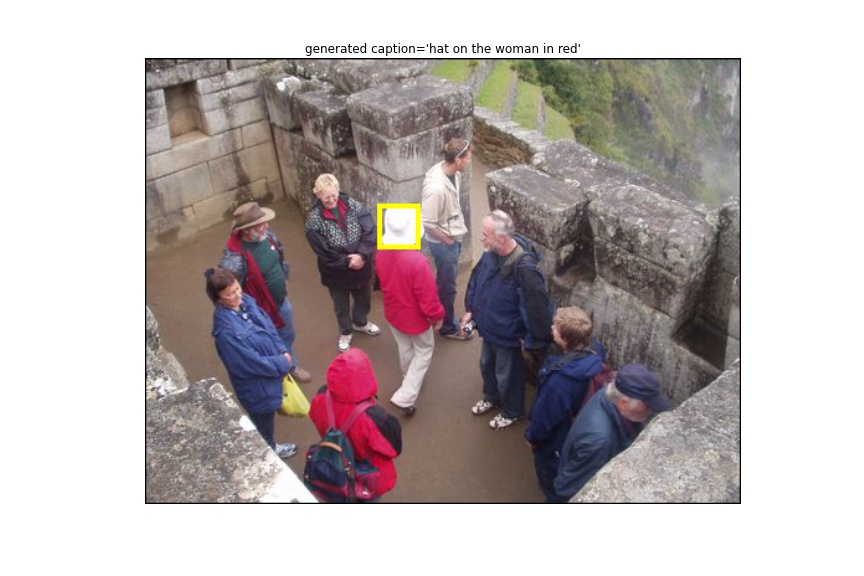} &
\includegraphics[trim = 40mm 25mm 40mm 20mm , clip=true,width=0.3\textwidth,height=\textheight,keepaspectratio]{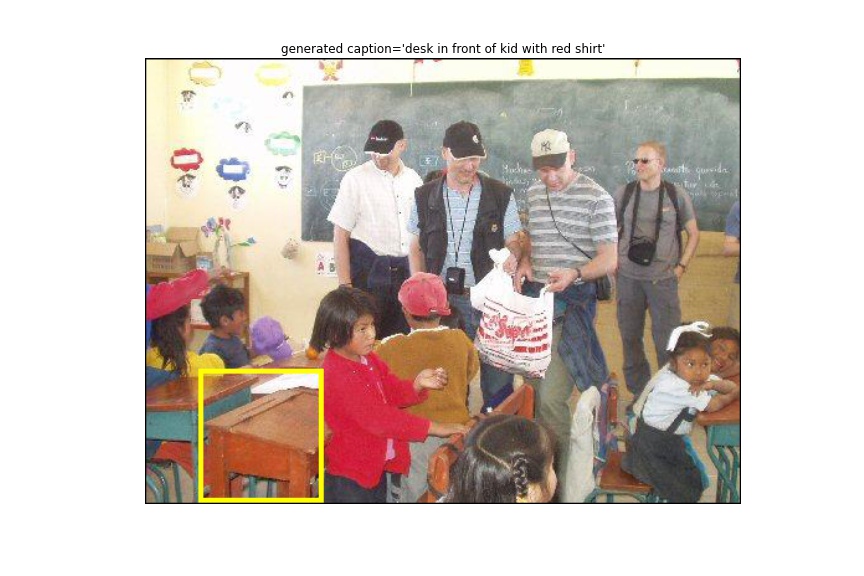} & 
\includegraphics[trim = 40mm 25mm 40mm 20mm , clip=true,width=0.3\textwidth,height=\textheight,keepaspectratio]{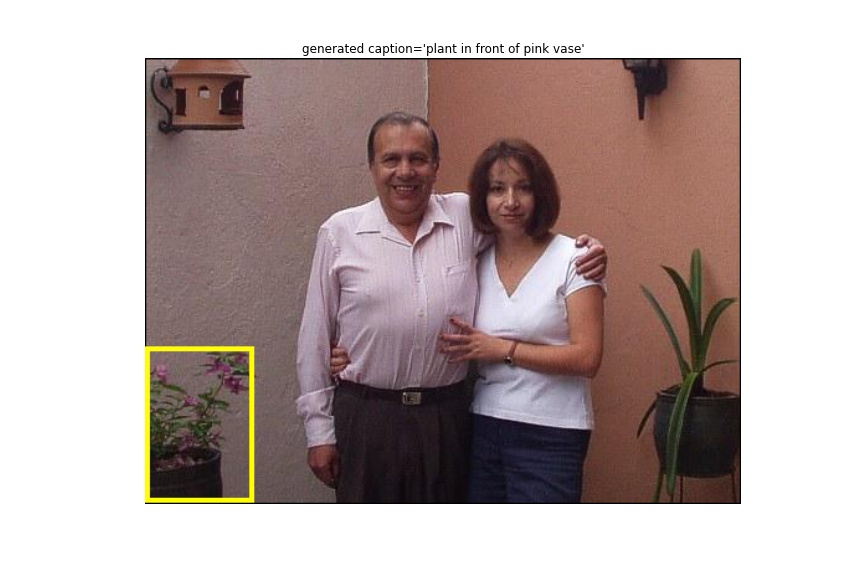} \\

{\small{generated description='\textit{sun}'}} &
{\small{generated description='\textit{sky}'}} &
{\small{generated description='\textit{tree trunk left}'}} \\
\includegraphics[trim = 40mm 25mm 40mm 20mm , clip=true,width=0.3\textwidth,height=\textheight,keepaspectratio]{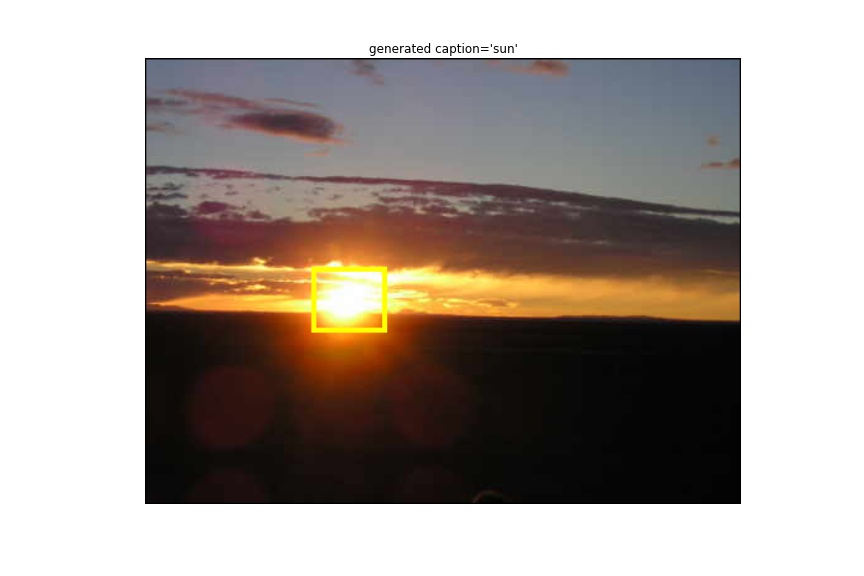} & 
\includegraphics[trim = 40mm 25mm 40mm 20mm , clip=true,width=0.3\textwidth,height=\textheight,keepaspectratio]{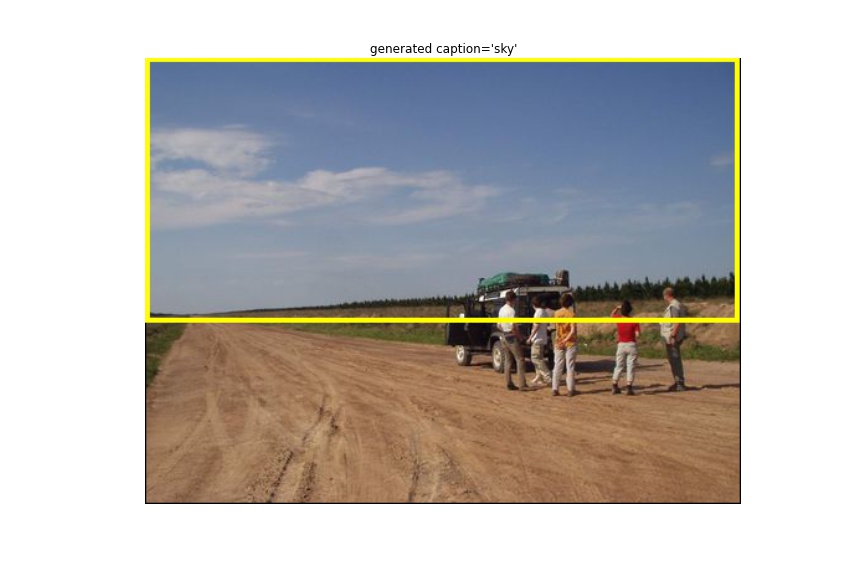} &
\includegraphics[trim = 40mm 25mm 40mm 20mm , clip=true,width=0.3\textwidth,height=\textheight,keepaspectratio]{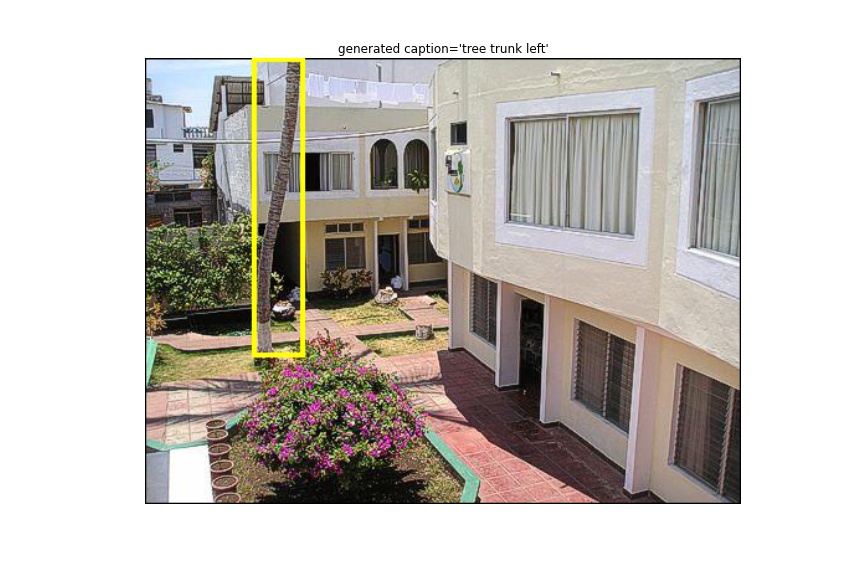} \\

\end{tabular}
\end{center}
\caption{Generated object descriptions by our model on ReferIt. The bounding box of the object is shown in yellow. }
\label{fig:referit_gencap}
\end{figure*}

\mysubsubsection{Generating descriptions for objects}
Although our SCRC model is designed for natural language object retrieval, it can also be applied in another task to generate descriptions for the objects in an image. Given an image $I_{im}$ and the bounding box of an object, a text description $S_{des}$ can be generated for the object as $S_{des} = \argmax_S p(S | I_{box}, I_{im}, x_{spatial})$ using beam search, where $I_{im}$ is the local image region of the object and $x_{spatial}$ is its spatial configuration. Figure \ref{fig:referit_gencap} shows some object descriptions generated by our SCRC model on ReferIt dataset.

\subsection{Object retrieval evaluation on Kitchen dataset}
\label{sec:exp_kitchen}

\begin{figure*}
\begin{center}
\begin{tabular}{c}
\small{query='\textit{whisk with red tipped handle}'}
\end{tabular}
\begin{tabular}{c@{}c@{}c@{}c@{}c@{}c@{}c@{}c@{}c@{}c@{}c}
\includegraphics[trim = 0mm 0mm 0mm 0mm, clip=true,width=0.15\textwidth, height=0.06\textheight,keepaspectratio]{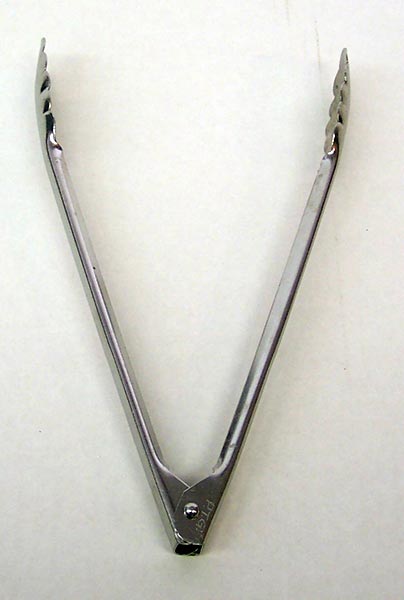} &
\includegraphics[trim = 0mm 0mm 0mm 0mm, clip=true,width=0.15\textwidth, height=0.06\textheight,keepaspectratio]{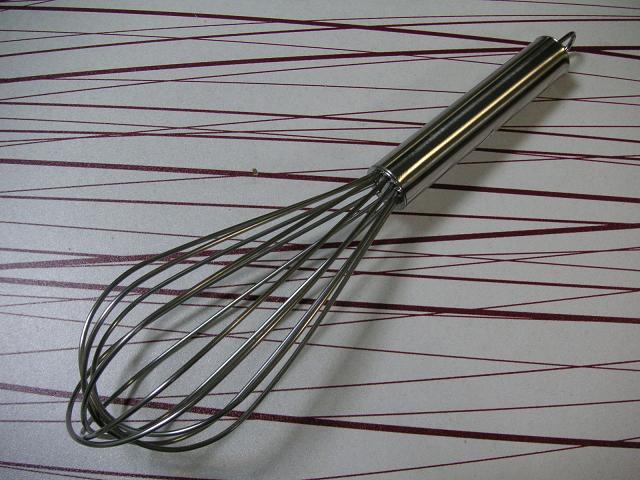} &
\includegraphics[trim = 0mm 0mm 0mm 0mm, clip=true,width=0.15\textwidth, height=0.06\textheight,keepaspectratio]{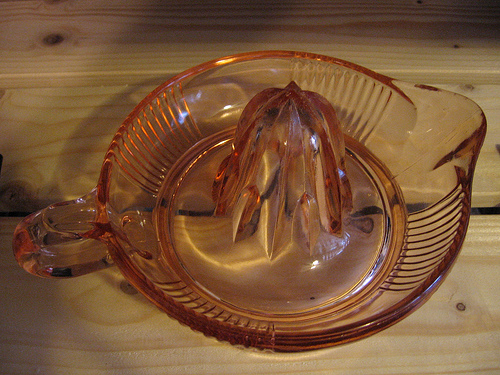} &
\includegraphics[trim = 0mm 0mm 0mm 0mm, clip=true,width=0.15\textwidth, height=0.06\textheight,keepaspectratio]{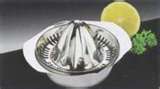} &
\includegraphics[trim = 0mm 0mm 0mm 0mm, clip=true,width=0.15\textwidth, height=0.06\textheight,keepaspectratio]{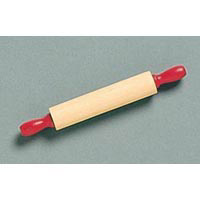} &
\includegraphics[trim = 0mm 0mm 0mm 0mm, clip=true,width=0.15\textwidth, height=0.06\textheight,keepaspectratio]{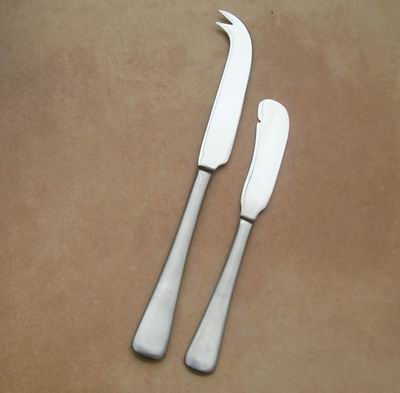} &
\includegraphics[trim = 0mm 0mm 0mm 0mm, clip=true,width=0.15\textwidth, height=0.06\textheight,keepaspectratio]{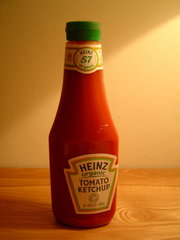} &
\includegraphics[trim = 0mm 0mm 0mm 0mm, clip=true,width=0.15\textwidth, height=0.06\textheight,keepaspectratio]{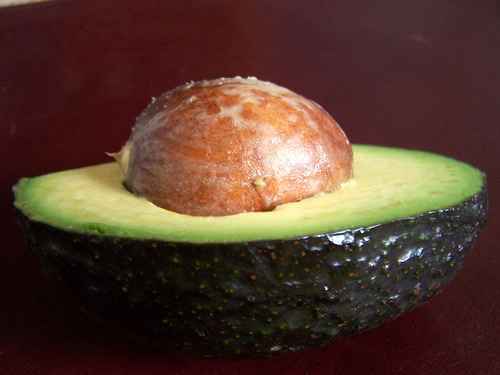} &
\includegraphics[trim = 0mm 0mm 0mm 0mm, clip=true,width=0.15\textwidth, height=0.06\textheight,keepaspectratio]{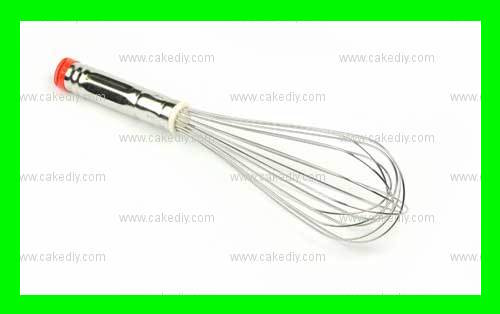} &
\includegraphics[trim = 0mm 0mm 0mm 0mm, clip=true,width=0.15\textwidth, height=0.06\textheight,keepaspectratio]{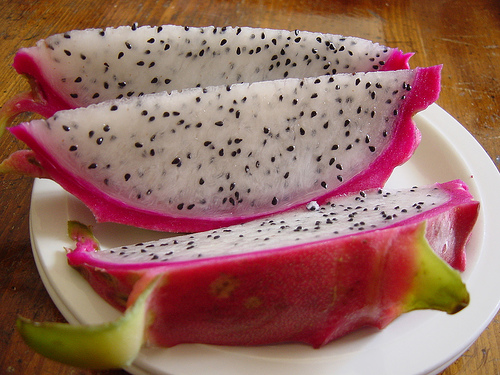} &
\includegraphics[trim = 0mm 0mm 0mm 0mm, clip=true,width=0.15\textwidth, height=0.06\textheight,keepaspectratio]{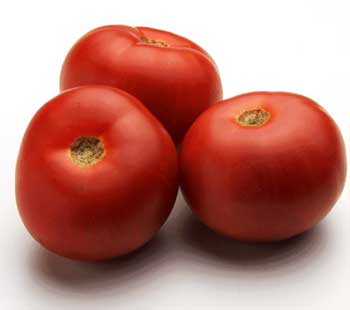} \\
\end{tabular}
\begin{tabular}{c}
\small{query='\textit{mobile phone the pink color}'}
\end{tabular}
\begin{tabular}{c@{}c@{}c@{}c@{}c@{}c@{}c@{}c@{}c@{}c@{}c}
\includegraphics[trim = 0mm 0mm 0mm 0mm, clip=true,width=0.15\textwidth, height=0.06\textheight,keepaspectratio]{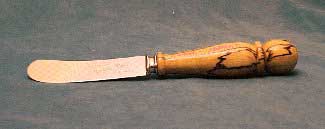} &
\includegraphics[trim = 0mm 0mm 0mm 0mm, clip=true,width=0.15\textwidth, height=0.06\textheight,keepaspectratio]{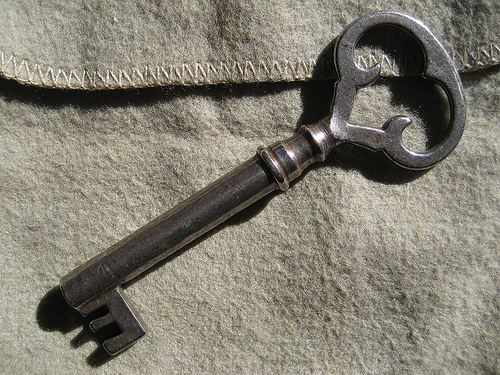} &
\includegraphics[trim = 0mm 0mm 0mm 0mm, clip=true,width=0.15\textwidth, height=0.06\textheight,keepaspectratio]{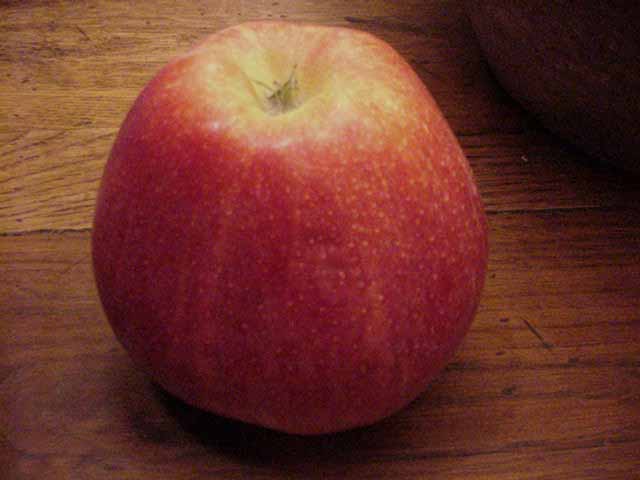} &
\includegraphics[trim = 0mm 0mm 0mm 0mm, clip=true,width=0.15\textwidth, height=0.06\textheight,keepaspectratio]{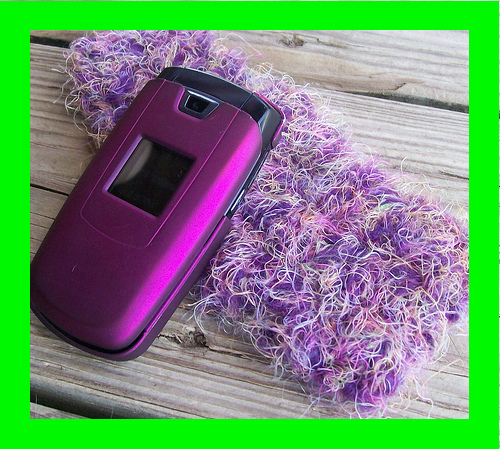} &
\includegraphics[trim = 0mm 0mm 0mm 0mm, clip=true,width=0.15\textwidth, height=0.06\textheight,keepaspectratio]{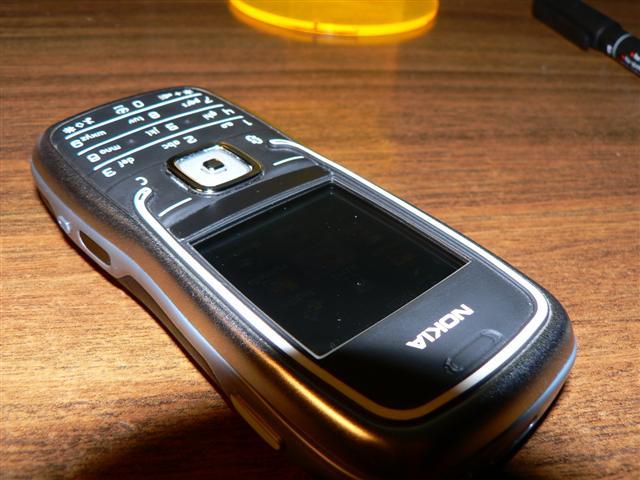} &
\includegraphics[trim = 0mm 0mm 0mm 0mm, clip=true,width=0.15\textwidth, height=0.06\textheight,keepaspectratio]{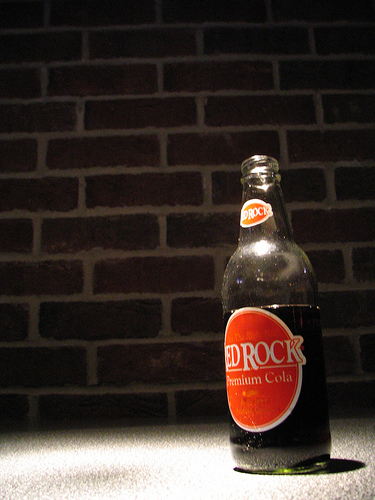} &
\includegraphics[trim = 0mm 0mm 0mm 0mm, clip=true,width=0.15\textwidth, height=0.06\textheight,keepaspectratio]{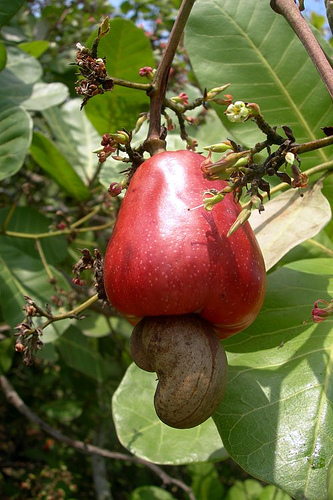} &
\includegraphics[trim = 0mm 0mm 0mm 0mm, clip=true,width=0.15\textwidth, height=0.06\textheight,keepaspectratio]{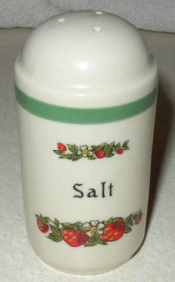} &
\includegraphics[trim = 0mm 0mm 0mm 0mm, clip=true,width=0.15\textwidth, height=0.06\textheight,keepaspectratio]{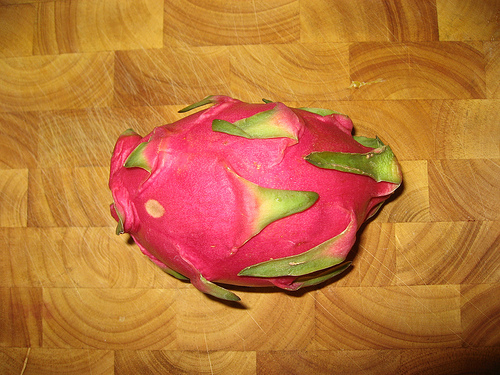} &
\includegraphics[trim = 0mm 0mm 0mm 0mm, clip=true,width=0.15\textwidth, height=0.06\textheight,keepaspectratio]{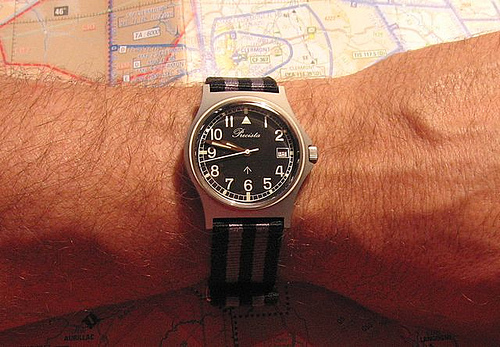} &
\includegraphics[trim = 0mm 0mm 0mm 0mm, clip=true,width=0.15\textwidth, height=0.06\textheight,keepaspectratio]{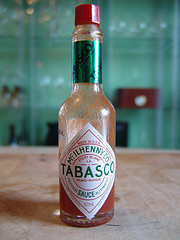} \\
\end{tabular}
\end{center}
\caption{Correctly retrieved examples in Kitchen dataset, where the highest scoring object (green) matches ground truth.}
\label{fig:kitchen_samples}
\end{figure*}

We also evaluate and compare our method with the baseline model \cite{guadarrama2014open} on the same Kitchen dataset as used in \cite{guadarrama2014open}. Kitchen is a dataset with 606 images sampled from the kitchen/household sub-tree of ImageNET hierarchy \cite{deng2009imagenet}, with 10 different descriptions annotated for each image. Since objects in this dataset almost occupy the entire images, instead of using retrievals on candidate object proposals boxes, in \cite{guadarrama2014open} the performance of the object retrieval is evaluated at image-level. During testing, for each query text, the candidate set consists of 11 images with  ground truth and 10 distractors. The distractors are sampled either from the same Kitchen dataset (``Kitchen'' experiment) or from the whole ImageNET (``ImageNET'' experiment), with the latter being an easier task. Performance of object retrieval is evaluated using top-1 precision.

To evaluate our method on this dataset, we split the dataset into two parts, with 300 images as trainval set and 306 images as test set. Similar to Section \ref{sec:exp_referit}, we first pretrain a SCRC model on MSCOCO dataset \cite{lin2014microsoft} for image captioning, and then fine-tune the model on the trainval set. The our model is tested through image-level retrieval on the candidate set of ground truth and 10 distractors, where we use the feature extracted from the entire image as $x_{box}$. Since the dataset involves no spatial configurations or scene-level contextual information, we set $x_{spatial}$ and $W_{global}$ in Eqn. \ref{eqn:next_word} to zero during fine-tuning and testing, so the model can only learn to score a candidate based $x_{box}$. As this dataset is a much smaller than ReferIt, we observe that transferring knowledge from MSCOCO significantly boosts the performance and avoids overfitting.

\begin{table}[t]
\begin{center}
\small
\begin{tabular}{|l|c|c|}
\hline
\textbf{Method} & Kitchen & ImageNet \\
\hline
CAFFE-7K \cite{guadarrama2014open}             &  51.34\%           & 57.50\%          \\
LRCN \cite{donahue2015long}                    &  40.35\%           & 63.22\%          \\
SCRC (w/o context, spatial, transfer)          &  54.02\%           & 74.08\%          \\
SCRC (w/o context, spatial)                    &  \textbf{61.62\%}  & \textbf{81.15\%} \\
\hline
\end{tabular}
\end{center}
\caption{Performance of different methods on the Kitchen dataset. See \secref{sec:kitchen:results} for details.}
\label{tab:results_kitchen}
\end{table}

\mysubsubsection{Results}
\label{sec:kitchen:results}
Table \ref{tab:results_kitchen} shows the top-1 precision (P@1) of our method together with the baseline on the test set. The first column ``Kitchen'' corresponds to sampling the 10 distractors from the same Kitchen dataset, while the second column corresponds to sampling distractors from the whole ImageNET 7K dataset \cite{deng2009imagenet}. Similar to Section \ref{sec:exp_referit}, \textbf{LRCN} refers to directly running LRCN model on the candidate images as a retriever. \textbf{SCRC (w/o context, spatial, transfer)} refers to the SCRC model directly trained on the trainval part of the Kitchen dataset, with convolutional layer initialized from VGG-16 net, and LSTM unit, word embedding and word prediction weights randomly initialized. \textbf{SCRC (w/o context, spatial)} corresponds to first pretraining on MSCOCO and then fine-tuning on Kitchen trainval set as described in Section \ref{sec:method}. As the dataset contains no spatial configuration or scene-level context information, we cannot test our full SCRC model on it. It can be seen from Table \ref{tab:results_kitchen} that in both scenarios, pretraining on image captioning and fine-tune on natural language object retrieval leads to the best performance, outperforming the baseline bag-of-words model CAFFE-7K and LRCN. Figure \ref{fig:kitchen_samples} shows some correctly retrieved object examples from Kitchen dataset, where the highest scoring candidate matches the ground truth. Both the ground truth and the 10 distractor images are sampled from the same Kitchen dataset in Figure \ref{fig:kitchen_samples}.

Moreover, as Kitchen dataset has only 606 objects and is more than 100 times smaller than ReferIt dataset, ``SCRC (w/o context, spatial)'' has significantly higher accuracy than ``SCRC (w/o context, spatial, transfer)''. This shows that pretraining on MSCOCO for image captioning dataset improves the performance of natural language object retrieval significantly on this relatively smaller dataset, by transferring the visual-linguistic knowledge from the former task to the latter task. As a reference, we note that \cite{guadarrama2014open} also uses an instance model and achieves higher overall performance. The instance model sends the query and candidate image regions to online APIs such as Google Image Search and FreeBase on the fly at test time. As in this work we assume a self-contained system that can be applied without resorting to Internet APIs on the fly, we only compare with the category model CAFFE-7K in \cite{guadarrama2014open}.

\subsection{Object retrieval evaluation on Flickr30K Entities dataset}

\begin{table}
\begin{center}
\small
\begin{tabular}{|l|c|c|}
\hline
\textbf{Method} & R@1 & R@10 \\
\hline
CCA \cite{plummer15iccv}                        &  25.3\%           &  59.7\%           \\
SCRC                                            &  \textbf{27.8\%}  &  \textbf{62.9\%} \\
\hline
Oracle                                          &  76.9\%           &  76.9\%           \\
\hline
\end{tabular}
\end{center}
\caption{Performance of our method compared with Canonical Correlation Analysis (CCA) baseline on 100 EdgeBox proposals in Flickr30K Entities dataset. Oracle corresponds to the highest possible recall on all 100 proposals for any retrieval method. }
\label{tab:results_flickr_edgebox}
\end{table}

We also train and evaluate our method on the Flickr30K Entities dataset \cite{plummer15iccv} for natural language object retrieval, which contains 31,783 images and 275,775 annotated bounding boxes. The object-level annotations in this dataset are derived from existing scene-level captions in Flickr30K \cite{young2014image}.

We train our model on the referential expressions in the Flickr30K dataset using the same top-100 EdgeBox \cite{zitnick2014edge} proposals same as in \cite{plummer15iccv}. On this dataset, our SCRC model achieves higher recall than the Canonical Correlation Analysis (CCA) method in \cite{plummer15iccv}, as is shown in Table \ref{tab:results_flickr_edgebox}.

\section{Conclusion}
\label{sec:conclude}

In this paper, we address natural language object retrieval with Spatial Context Recurrent ConvNet (SCRC), a recurrent neural network model that scores a candidate box based on local image descriptors, spatial configurations and global scene-level context. We show that incorporation of spatial configuration and global context improves the performance of natural language object retrieval significantly. The recurrent network model used in our method leads to an end-to-end trainable scoring function, which significantly outperforms baseline methods.

Also, we demonstrate that natural language object retrieval can benefit from transferring knowledge learned on image captioning through pretraining and adaptation. As one of the difficulties for natural language object retrieval systems is the lack of large datasets with object-level annotation, we show that this problem can be alleviated by exploiting datasets with image-level annotations, which are often easier to collect than object-level descriptions. As follow up to this work we show successful results by encoding the phrase rather than scoring it \cite{rohrbach2015grounding} and also predicting image segmentations instead of bounding boxes \cite{hu2016segmentation}.

\section*{Acknowledgments}
The authors are grateful to Lisa Hendricks, Jeff Donahue, Subhashini Venugopalan and Coline Devin for helpful feedback on drafts.
M. Rohrbach was supported by a fellowship within the FITweltweit-Program of the German Academic Exchange Service (DAAD). J. Feng was supported by NUS startup grant R‐263‐000‐C08‐133. This work was supported by DARPA, AFRL, DoD MURI award N000141110688, NSF awards IIS-1427425 and IIS-1212798, and the Berkeley Vision and Learning Center.

\small
\bibliographystyle{ieee}
\bibliography{biblioLong,references_arxiv}

\end{document}